\documentclass[pdflatex,sn-nature]{sn-jnl}

\usepackage{graphicx}%
\usepackage{multirow}%
\usepackage{amsmath,amssymb,amsfonts}%
\usepackage{amsthm}%
\usepackage{mathrsfs}%
\usepackage[title]{appendix}%
\usepackage{xcolor}%
\usepackage{textcomp}%
\usepackage{manyfoot}%
\usepackage{booktabs}%
\usepackage{algorithm}%
\usepackage{algorithmicx}%
\usepackage{algpseudocode}%
\usepackage{listings}%
\usepackage{subcaption}
\usepackage{caption}
\usepackage{placeins}
\usepackage{afterpage,placeins}

\usepackage{lineno}
\usepackage{geometry}
\usepackage[T1]{fontenc}

\theoremstyle{thmstyleone}%
\theoremstyle{thmstyletwo}%

\theoremstyle{thmstylethree}%

\newcommand{\torevise}[1]{{\color{black}#1}}

\begin{document}

\title[Article Title]{MicroscopyMatching: Towards a Ready-to-use Framework for Microscopy Image Analysis in Diverse Conditions}

\author[1]{\fnm{Xiaofei} \sur{Hui}}\email{x.hui@lancaster.ac.uk}
\equalcont{These authors contributed equally to this work.}

\author[1]{\fnm{Haoxuan} \sur{Qu}}\email{h.qu5@lancaster.ac.uk}
\equalcont{These authors contributed equally to this work.}

\author[1]{\fnm{Hossein} \sur{Rahmani}}\email{h.rahmani@lancaster.ac.uk}

\author[2]{\fnm{Shuohong} \sur{Wang}}\email{wangsh@hms.harvard.edu}

\author[3]{\fnm{Jeff W.} \sur{Lichtman}}\email{jeff@mcb.harvard.edu}

\author*[1]{\fnm{Jun} \sur{Liu}}\email{j.liu81@lancaster.ac.uk}

\affil[1]{\orgdiv{School of Computing and Communications}, \orgname{Lancaster University}, \orgaddress{\country{Lancaster, LA1 4WA, UK}}}

\affil[2]{\orgdiv{Department of Cell Biology}, \orgname{Harvard Medical School}, \orgaddress{Boston, MA 02115, USA}}

\affil[3]{\orgdiv{Department of Molecular and Cellular Biology, Center for Brain Science}, \orgname{Harvard University}, \orgaddress{\country{Cambridge, MA 02138, USA}}}

\abstract{
Analyzing microscopy images to extract biological object properties (e.g., their morphological organization, temporal dynamics, and population density) is fundamental to various biomedical research. Yet conducting this manually is costly and time-consuming. Though deep learning-based approaches have been explored to automate this process, the substantial diversity of microscopy analysis settings in practice (including variations of biological object types, sample processing protocols, imaging equipment, and analysis tasks, etc.) often renders them ineffective. As a result, these approaches typically require extensive adaptation for different settings, which, however, can impose burdens that are often practically unsustainable for laboratories, forcing biomedical researchers to still commonly rely on manual analysis, thereby severely bottlenecking the pace of biomedical research progress.
This situation has created a pressing and long-standing need for a reliable and broadly applicable microscopy image analysis tool, yet such a tool is still missing. To address this gap, we present the first ready-to-use microscopy image analysis framework, MicroscopyMatching, that can reliably perform key analysis tasks (including segmentation, tracking, and counting) across diverse microscopy analysis settings. From a fundamentally different perspective, MicroscopyMatching reformulates diverse microscopy image analysis tasks as a unified matching problem, effectively handling this problem by exploiting the robust matching capability from pre-trained latent diffusion models. Through this distinctive design, MicroscopyMatching achieves consistently superior performance across 20 benchmark datasets and a collection of image data from 200 sets of diverse, real-world experiments, with performance even comparable to human experts, in contrast to existing methods frequently failing to provide reliable results. Together, these results establish MicroscopyMatching as a practical and ready-to-use tool, which shall bring substantial benefit to various biomedical research in real-world settings.
}

\maketitle

\section*{Introduction}\label{sec:intro}

Microscopy is a central technology in modern biomedical research \cite{caicedo2019nucleus}, with more than 72\% of life-science laboratories relying on at least one advanced microscopy platform \cite{360researchreportsLifeScience} and over one million PubMed-indexed scientific publications related to microscopy to date \cite{caicedo2019nucleus,nihMicroscopyMicroscope}. In this context, analyzing microscopy images to extract properties (e.g., morphological organization, temporal dynamics, and population density) of biological objects has become a fundamental component of biomedical research \cite{caicedo2019nucleus,falk2019u,mavska2014benchmark}. These analyses support a broad range of biomedical investigations, such as profiling the morphology and spatial organization of tumor cells in complex tissues \cite{romero2024neuroendocrine}, monitoring cell dynamics during cancer progression \cite{panagopoulos2025multigenerational}, and quantifying immune cell populations in tissue microenvironments \cite{chi2025dietary}. However, microscopy images often contain large numbers of densely packed biological objects, making it particularly burdensome and challenging for manual extraction of biological object properties, such as manually segmenting, tracking (tracing), or counting target biological objects.
This burden is further amplified by the scale and prevalence of microscopy-based experiments: as a single experiment may generate thousands of images \cite{boutros2015microscopy}, and such experiments are conducted widely across biomedical research studies, manual analysis becomes extremely labor-intensive and time-consuming, which could even divert researchers' time and effort from core scientific activities, ultimately hindering overall research progress.

\begin{figure}[ht]
    \centering
    \includegraphics[width=0.9\linewidth]{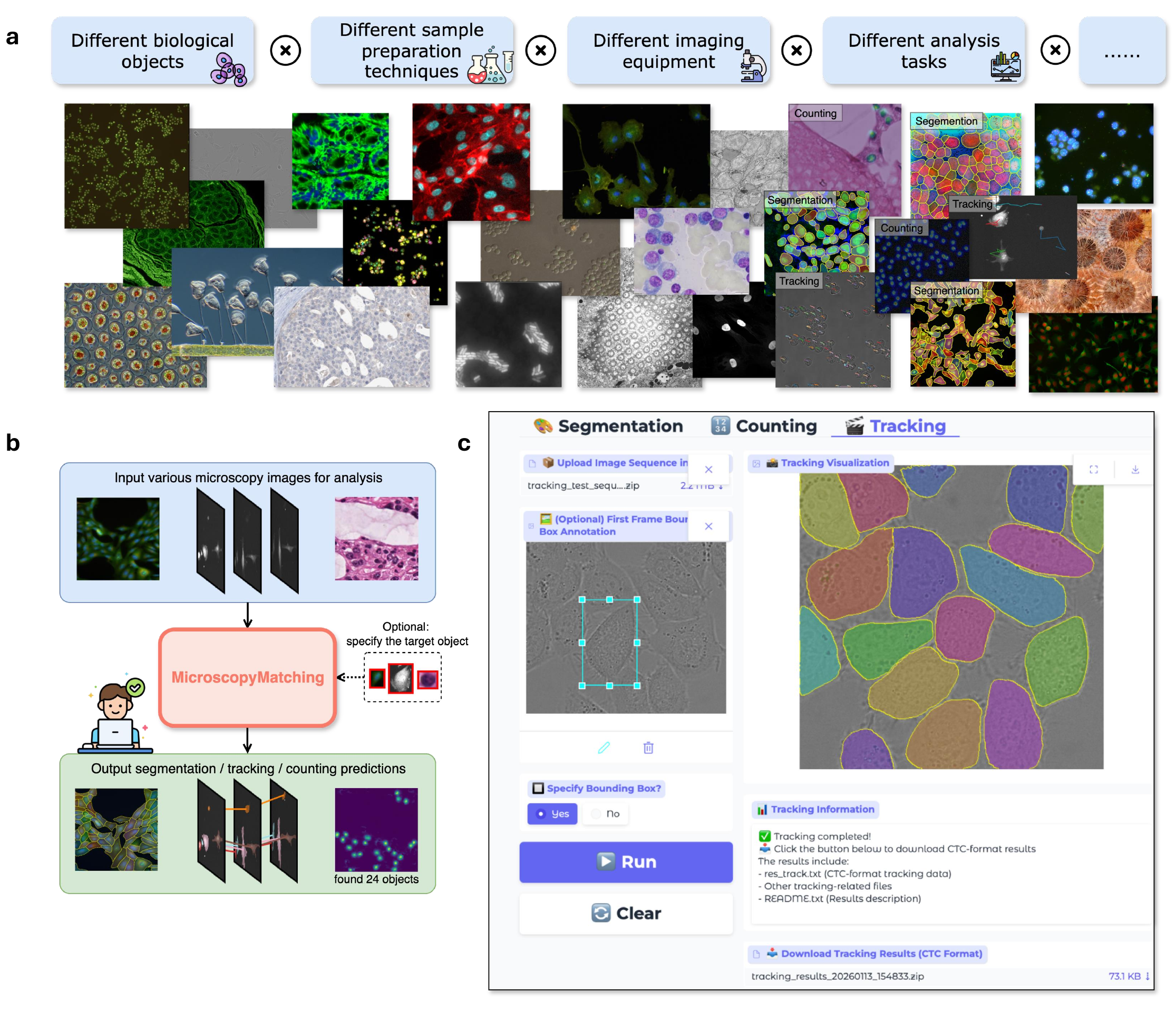}
    \caption{\textbf{Overview of MicroscopyMatching.} 
    \textbf{(a)}, 
    Real-world microscopy image analysis spans a wide range of experimental settings and analytical requirements. Variations arise from differences in biological objects (e.g., distinct cell lines or specific tissue samples), sample preparation protocols (e.g., using different staining techniques to highlight specific proteins), microscopy equipment (e.g., brightfield microscopy, fluorescence microscopy, transmission electron microscopy, etc.), and analysis tasks (e.g., segmentation, tracking, and counting), resulting in highly diverse image appearances and complex analysis scenarios. 
    \textbf{(b)},
    Our MicroscopyMatching framework performs microscopy image analysis by conducting matching within the input image data. The matching is designed to be flexible in usage: Users may optionally provide an exemplar bounding box specifying the target object in the image, enabling the framework to perform matching between the exemplar and other regions of the image; Alternatively, users can also let the framework automatically perform matching between repetitive objects in the image data without providing exemplars.
    Based on the matching mechanisms, MicroscopyMatching reliably performs segmentation, tracking, and counting within a single framework. 
    \textbf{(c)}, Screenshot of the MicroscopyMatching tool interface, taking the tracking task as an example (see screenshots for segmentation and counting tasks in Supplementary Figure 1). 
    When using the MicroscopyMatching tool, users only need to select an analysis task (segmentation, tracking, or counting; tracking is selected in this figure), upload the test image data, optionally specify a target object by providing a bounding-box exemplar in the image (for segmentation and counting tasks) or in the first frame of the image sequence (for the tracking task), and click ``Run'' to perform the analysis.   
    Analysis results are displayed on the right side of the tool interface, including visualization of the prediction results and downloadable predictions in standard formats.
    The tool is publicly available at \url{https://huggingface.co/spaces/VisionLanguageGroup/MicroscopyMatching}.}
    \label{fig:overview}
\end{figure}

To remedy this, various deep learning methods ~\cite{pang2025cellotype,archit2025segment,edlund2021livecell,greenwald2022whole,ershov2022trackmate,zheng2024rethinking} have been proposed to automate microscopy image analysis. Existing conventional deep learning models \cite{pang2025cellotype,edlund2021livecell,greenwald2022whole,conrad2023instance,spahn2022deepbacs} are generally dedicated to specific microscopy image analysis settings (e.g., particular combinations of microscopy equipment and biological objects). Through data-intensive and setting-specific training, they can achieve good performance when the test data closely matches their training settings \cite{greenwald2022whole,edlund2021livecell,stringer2025cellpose3,akers2021rapid,ershov2022trackmate,zheng2024rethinking}. More recently, foundation models such as SAM~\cite{kirillov2023segment,carion2025sam} and CLIP~\cite{radford2021learning}, originally pre-trained for general vision domains, have also motivated several efforts to adapt them to biomedical applications via fine-tuning to achieve better results~\cite{archit2025segment,ma2024segment,jin2024sem,israel2025cellsam,pachitariu2025cellpose}. However, despite the advances, a critical gap remains: All these existing methods, including foundation model adaptations, still struggle (and often fail) to provide reliable outputs in real-world microscopy research workflows. This is because, in practice, microscopy image analysis is typically applied across highly diverse and continually evolving settings, where variations in settings often translate into substantial domain shifts \cite{zhou2023domain} that degrade the performance of existing methods. More specifically, biomedical research across different laboratories (and even within the same laboratory) may target different biological objects, focus on distinct biological features, or examine specific biological responses, each requiring tailored sample processing protocols and specialized microscopy equipment. Moreover, as biomedical research continues to advance, new biological objects, new sample processing protocols, and even new microscopy equipment are continuously introduced to accommodate emerging research needs \cite{sakthivelu2025functional,han2024multiplexed,wu2024mapping}. 
This leads to an ever-growing variability in experimental settings and corresponding microscopy image characteristics (Fig.~\ref{fig:overview}a), continually shifting the underlying data distribution and inducing new domain shifts. In addition to this variability in image characteristics, different biomedical research can also impose distinct analytical needs, requiring different analysis tasks (such as segmentation \cite{stringer2021cellpose}, tracking \cite{ershov2022trackmate}, and counting \cite{zheng2024rethinking}) to be performed to extract different biological properties. Collectively, the diversity, spanning over different (and even new) biological objects, sample preparation protocols, microscopy equipment, and analysis tasks, creates a fundamentally heterogeneous and dynamically evolving landscape for microscopy image analysis in practice, which poses a significant challenge to existing methods.

In this context, to cope with the heterogeneity and the resulting domain shifts, existing methods often require specific and careful model adaptation for different settings to obtain reliable analysis results. This process includes producing sufficient and high-quality annotations for optimization, and meticulously tuning the model's parameters, which, however, creates a prohibitive and often even impractical barrier for biomedical researchers. This is because: (1) Producing high-quality annotations for model adaptation and optimization is already labor-intensive for a single setting, often requiring hundreds of hours of expert labor to annotate even a single dataset \cite{ker2018phase}. Far worse, as microscopy settings vary widely and continually evolve, this annotation effort can be repeatedly required for new settings, creating a compounding burden that severely bottlenecks the pace of biomedical research discovery. (2) Even with annotations prepared, training and adapting models still often fail to produce satisfactory results \cite{varoquaux2022machine}, necessitating repeated troubleshooting and hyperparameter tuning that often requires repeated back-and-forth with deep-learning engineers, leading to high operational costs and long turnaround time. Since such adaptation efforts recur for different settings, the cumulative demands in costs and time can quickly become untenable in practice for many laboratories. As a result, biomedical researchers in practice still often rely on manual analysis to obtain reliable results \cite{sakthivelu2025functional,han2024multiplexed,shapson2021connectomic}. This situation has thus led to a long-standing and pressing need for a ``ready-to-use'' microscopy image analysis tool: one that works robustly across diverse microscopy analysis settings without additional training data annotation or machine learning model tuning, while consistently delivering reliable analysis results. However, such a tool remains largely unavailable to biomedical researchers.

\FloatBarrier

To fill this critical gap, in this paper, we aim to develop such a practically ready-to-use tool that can reliably operate across diverse experimental settings on key analysis tasks (including segmentation, tracking, and counting). To achieve this, we first reconsider: what makes existing methods so sensitive to domain shifts? We identify that the root cause lies in how these tasks are fundamentally formulated: Existing methods typically formulate microscopy image analysis tasks as high-level object understanding problems, such as recognizing specific biological objects, interpreting their morphological features, and distinguishing them from background across varying imaging conditions. Such high-level understandings are inherently domain-specific \cite{stacke2020measuring}, as the learned features and representations are tied to particular experimental settings (specific cell types, sample preparations, and imaging equipment), thereby leading to performance degradation and adaptation requirements when settings change.
This limitation thus motivates us to reformulate the tasks from a fundamentally different perspective. To this end, we observe that, despite the diversity in microscopy image analysis settings, segmentation, tracking, and counting can all be cast as matching problems: As biological objects in microscopy images typically appear densely with similar and repetitive visual patterns, these objects can be identified by ``within-image'' matching, either by specifying an exemplar within the image to indicate the analytical target and matching the exemplar to the rest of the image, or by exploiting these repetitive patterns to infer the underlying object structure and perform mutual matching among the objects.
Under this reformulation, segmentation can be cast as delineating matched regions in the image, tracking can be viewed as identifying matched regions in each frame and matching them across successive frames, and counting can be performed by tallying all matched instances in the image. Critically, this matching-based formulation fundamentally simplifies the problem. Rather than requiring high-level understandings of biological objects across varying conditions, the matching-based formulation relies on a more basic, lower-level visual comparison: As the comparison (whether between an exemplar and target objects or among similar objects within the image) occurs within the same image (or image sequence) and therefore lies within the same imaging domain, the problem essentially reduces to ``within-domain'' low-level comparison, which is substantially simpler than learning to recognize objects across diverse domains. This inherent simplicity makes matching fundamentally robust to domain variations and therefore feasible to develop broadly-applicable tools that can reliably operate across diverse experimental settings. Hence, from this perspective, the seemingly challenging problem of handling diverse microscopy analysis tasks across heterogeneous conditions can be abstracted into a single core challenge: robust matching. This then leaves us with a key question: how can such robust matching be achieved effectively?

To answer this question, we draw inspiration from latent diffusion models (LDMs) \cite{rombach2022high}, a class of large foundational generative models originally developed for high-quality image generation. We identify that LDMs possess two key properties that make them particularly suitable for robust matching in microscopy images. First, being able to accurately generate images with very rich fine-grained details suggests that LDMs can capture dense visual structure, including local textures, object boundaries, shapes, and spatial layouts, which is highly beneficial for performing matching in microscopy images. Second, beyond image synthesis capability, the internal self- and cross-attention in LDMs also provide natural mechanisms for fine-grained matching. Specifically, during training, the self- and cross-attention layers are explicitly optimized to establish visual correspondences within an image context, associating visually consistent regions based on similarity under the same imaging conditions. This leads LDMs to acquire a fundamental capability of ``within-domain'' matching: identifying image regions that match a given image patch under the same imaging conditions. Through extremely large-scale training on billions of diverse images, this fundamental capability becomes highly robust. As observed in our previous works \cite{hui2024class,peng_harnessing}, such LDMs can perform robust matching for images with diverse conditions, including those not encountered during training.

The above observations lead to the insight that, if the fundamental fine-grained matching capability of LDMs can be effectively extracted and leveraged, robust microscopy image analysis can be achieved across diverse and evolving analysis needs. Building on this insight, we present MicroscopyMatching, a novel framework that provides the first ready-to-use tool to reliably perform microscopy image analysis across highly heterogeneous experimental settings and key tasks (including segmentation, tracking, and counting) in practice. It is noteworthy that MicroscopyMatching requires no burdensome effort of model adaptation or fine-tuning for each new experimental setting. Instead, it enables flexible and user-friendly microscopy image analysis through ``within-domain'' matching. Specifically, after uploading microscopy image data (image or image sequence) to MicroscopyMatching, users can either specify a bounding box containing an exemplar of the target objects in the image data (Fig.~\ref{fig:overview}c), allowing the model to perform matching between the exemplar and other regions in the uploaded data, or simply let the model automatically discover and match visually recurring structures within the same image data without providing any exemplar box. These features together shall free biomedical researchers from the substantial effort of data annotation and model adaptation and tuning, making MicroscopyMatching the first highly scalable, broadly applicable, and ready-to-use solution for microscopy image analysis.

Compared to existing methods that frequently fail in practice under diverse and unseen experimental settings, we observe that MicroscopyMatching consistently delivers reliable results, with performance even comparable to that of humans. More specifically, we conducted the following multi-layered evaluations: (1) We benchmarked MicroscopyMatching against state-of-the-art methods for microscopy image segmentation, tracking, and counting across twenty publicly available benchmark datasets, where MicroscopyMatching consistently achieved superior performance (Fig. \ref{fig:result1} and Fig. \ref{fig:results2}). (2) Beyond standard benchmarks, to more specifically reflect the real-world variability that researchers encounter in practice, we also assembled a vast collection of heterogeneous microscopy image data, where each example was captured from a different microscopy image analysis setting, involving microscopy image data from around 135 laboratories and 200 sets of experiments. We observe that existing methods drastically fail under this real-world heterogeneity (Fig. \ref{fig:results3}), while MicroscopyMatching still consistently achieves strong performance across all tasks and experimental scenarios. (3) Having demonstrated its good performance, we further assessed whether MicroscopyMatching achieves reliability at a practically meaningful level through expert-rating and human-model agreement analyses, which demonstrate that MicroscopyMatching achieves human-level performance (Fig. \ref{fig:human_model}). (4) A user study involving biomedical professionals from diverse domains further shows that our tool can be readily adopted in practical use cases (Fig. \ref{fig:user_study}).
Taken together, these evaluations show that MicroscopyMatching is a high-performing solution across diverse and even previously unseen microscopy conditions. MicroscopyMatching shall benefit the research community by providing a reliable and accessible tool for microscopy image analysis, significantly alleviating the practical burden researchers face in their daily workflows. The code is released (at \url{https://github.com/phoebehxf/MicroscopyMatching}) to facilitate local deployment. We also provide an online interface for public demonstration at \url{https://huggingface.co/spaces/VisionLanguageGroup/MicroscopyMatching}.

\section*{Results}\label{sec:results}

\subsection*{MicroscopyMatching Formulation and Validation}

MicroscopyMatching aims to reliably perform key microscopy image analysis tasks across diverse experimental settings. It achieves this by casting these diverse tasks under heterogeneous settings into a unified matching-based problem, and leveraging the robust fine-grained matching capability of pre-trained LDM to effectively tackle this problem. 
We describe the MicroscopyMatching framework below using its application to segmentation as an example, as illustrated in Fig.~\ref{fig:result1}a. Further details of MicroscopyMatching are provided in Methods.

Given an input image, MicroscopyMatching first extracts latent features from the image using a variational autoencoder (VAE) module \cite{kingma2013auto}. It then derives an embedding feature and performs matching in two different usage modes, depending on whether an exemplar bounding box is provided by the user. When no exemplar bounding box is provided (the default mode, denoted as ``MicroscopyMatching\_A''), MicroscopyMatching obtains an internal embedding for identifying repetitive patterns in the input image (see Methods for details), which subsequently triggers the diffusion module \cite{rombach2022high} to perform automatic matching within the input image. When an exemplar bounding box is additionally provided (denoted as ``MicroscopyMatching\_S''), MicroscopyMatching performs matching between the exemplar and image regions by first projecting the exemplar into a compact embedding and then matching this exemplar embedding with the latent features in the diffusion module. In both modes, the matching process produces attention maps that provide direct insight into the target (auto-determined or user-specified) biological objects. Finally, the attention maps are fed into an attention post-processing module to generate the final output of the task, such as segmentation masks in the case of the segmentation task. Besides segmentation, MicroscopyMatching employs a similar procedure for tracking and counting (see more details in Methods). Notably, in the MicroscopyMatching framework, the VAE and the diffusion module are adopted from the pre-trained LDM and are kept fixed, while the lightweight attention post-processing modules and exemplar projector are trainable components. Note that the attention post-processing modules are shared across both usage modes, and the exemplar projector is only adopted in the MicroscopyMatching\_S mode.

Having established the framework above, we then describe the training and evaluation protocols of MicroscopyMatching (see Methods for more details). We first introduce the protocol under MicroscopyMatching's default usage mode, MicroscopyMatching\_A. For the segmentation task, we compared MicroscopyMatching with four state-of-the-art microscopy image segmentation methods (i.e., Cellpose3 \cite{stringer2025cellpose3}, CelloType \cite{pang2025cellotype}, MicroSAM \cite{archit2025segment}, and MitoNet \cite{conrad2023instance}), with all methods trained on the training sets of four commonly-used microscopy segmentation datasets: Cellpose Cyto \cite{stringer2021cellpose}, TissueNet \cite{greenwald2022whole}, LiveCell \cite{edlund2021livecell}, and Lucchi \cite{lucchi2012structured}. For tracking, we compared MicroscopyMatching with three current microscopy tracking pipelines (TrackMate \cite{ershov2022trackmate}, Trackastra \cite{gallusser2024trackastra}, and Ultrack \cite{bragantini2025ultrack}), with training data from four microscopy tracking datasets in the Cell Tracking Challenge \cite{mavska2023cell}: BF-C2DL-MuSC, BF-C2DL-HSC, Fluo-C2DL-MSC, and PhC-C2DH-U373. For counting, we compared MicroscopyMatching with two current microscopy counting methods (U-Net \cite{falk2019u} and DCL \cite{zheng2024rethinking}), all trained on the training sets of two counting datasets: DCC \cite{marsden2018people} and VGG \cite{lempitsky2010learning}.
The training protocol described

\begin{figure}[H]
    \centering
    \includegraphics[width=0.85\linewidth]{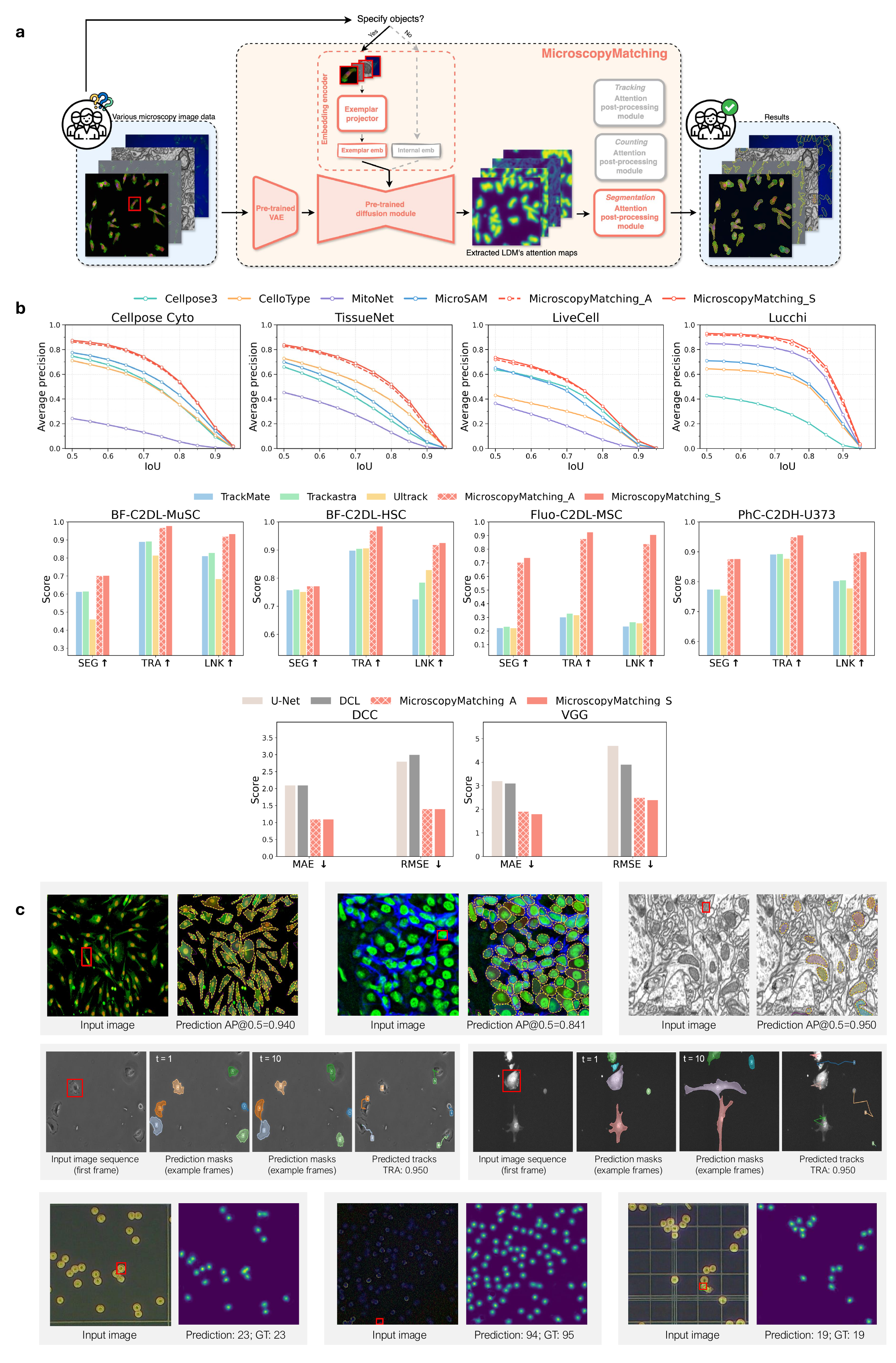}
\end{figure}
\FloatBarrier
\afterpage{
\captionsetup{type=figure}
\captionof{figure}{
    Framework formulation and validation for MicroscopyMatching.  \textbf{(a)}, Overview of MicroscopyMatching framework architecture, using its application to segmentation as an example. \textbf{(b)}, Evaluation of MicroscopyMatching\_A and MicroscopyMatching\_S on segmentation, tracking, and counting. 
    For each testing sample, MicroscopyMatching\_A receives only the input image data, and MicroscopyMatching\_S receives the input image data along with an exemplar bounding box. For segmentation, the line plots show AP scores across the intersection over union (IoU) thresholds from 0.5 to 0.95 on the evaluation set of each dataset (higher AP score indicates better segmentation performance). For tracking, we present bar plots showing SEG, TRA, and LNK scores on the evaluation set of each dataset (where higher scores indicate better tracking performance). For counting, we present bar plots showing MAE and RMSE scores on the evaluation set of each dataset (where lower MAE and RMSE scores indicate lower counting errors and better performance). See Methods for more details about the datasets and metrics. \textbf{(c)}, Qualitative results of MicroscopyMatching\_S on segmentation, tracking, and counting (qualitative results of MicroscopyMatching\_A are presented in Supplementary Figure 2). For segmentation, we show the input image with the exemplar bounding box in red, along with the predicted segmentation (in yellow dashed lines) and the ground-truth segmentation (in purple solid lines) overlaid on the input image. The image-level AP score at IoU threshold of 0.5 (AP@0.5) is shown below each example. For tracking, we show the first frame of the input image sequence (with the exemplar bounding box in red), predicted instance masks labeled with instance IDs across example frames (the corresponding frame index $t$ is marked on the top left corner of the frames), and the resulting cell trajectories obtained by linking the centers of the segmented instances across frames. The TRA score of the prediction is shown below the predicted trajectory of each example. For counting, we show the input image with the exemplar bounding box in red, along with the predicted density map. The density map represents how densely objects are distributed across the image, and the sum of values in the predicted density map gives the total predicted object count. The corresponding predicted and ground-truth (GT) object counts are shown below each example.}
\label{fig:result1}
\vspace{2mm}}

\noindent above applies to the shared components in both MicroscopyMatching\_A and MicroscopyMatching\_S modes.
For the MicroscopyMatching\_S mode, we additionally train the exemplar projector using exemplar-target pairs (more details in Methods).

After training, we evaluated MicroscopyMatching on the held-out evaluation sets corresponding to the above datasets, and compared with the existing methods described above (Fig. \ref{fig:result1}b and Supplementary Tables 1–3).
We evaluated MicroscopyMatching under both usage modes in each experiment.
The segmentation performance is reported using Average Precision (AP) metric following \cite{stringer2021cellpose,stringer2025cellpose3,pachitariu2025cellpose}. The tracking performance is reported using metrics from the Cell Tracking Challenge \cite{mavska2023cell}, including segmentation accuracy measure (SEG), tracking accuracy measure (TRA), and linking accuracy measure (LNK). The counting performance is reported using mean absolute error (MAE) and root mean square error (RMSE) metrics following \cite{sam2017switching,zheng2024rethinking} (see Methods for details of the metrics).
As shown in Fig.~\ref{fig:result1}b, across different tasks, MicroscopyMatching (in both usage modes) consistently and significantly outperformed all compared methods. We also present qualitative results of MicroscopyMatching for individual images (or image sequences) in Fig.~\ref{fig:result1}c and Supplementary Figure 2.

After constructing the MicroscopyMatching framework and observing its strong performance on the held-out evaluation sets, we next turn to the core question in this work: can MicroscopyMatching maintain robust performance and be genuinely ready-to-use in practical microscopy image analysis without additional fine-tuning, where imaging conditions are highly diverse and often previously unseen? We examined this question across four levels of evaluation that together reflect MicroscopyMatching's practical readiness: (1) generalization on unseen benchmarks, (2) robustness under real-world experimental heterogeneity, (3) reliability compared to human annotators, (4) practical usability and readiness for biomedical researchers. Importantly, in all the following evaluations and experiments, MicroscopyMatching was applied as-is, without any additional fine-tuning or adaptation after its construction.

\subsection*{Evaluation on Unseen Benchmark Datasets}

To evaluate MicroscopyMatching under unseen microscopy image analysis settings, we assessed it on a suite of ten diverse benchmark datasets that were not used for training. These cover segmentation datasets (NeurIPS CellSeg \cite{ma2024multimodality}, Xenium \cite{pang2025cellotype}, DeepBacs \cite{spahn2022deepbacs}, TEM Bench \cite{conrad2023instance}), tracking datasets (DIC-C2DH-HeLa, Fluo-N2DH-GOW1, Fluo-N2DL-HeLa, Fluo-N2DH-SIM+ from the Cell Tracking Challenge \cite{mavska2023cell}), and counting datasets (ADI \cite{paul2017count}, MBM \cite{kainz2015you}). Collectively, they span diverse unseen variations of microscopy experiment settings, providing a comprehensive benchmark evaluation of generalization ability (see Methods for more details).

\begin{figure}[H]
    \centering
    \includegraphics[width=\linewidth]{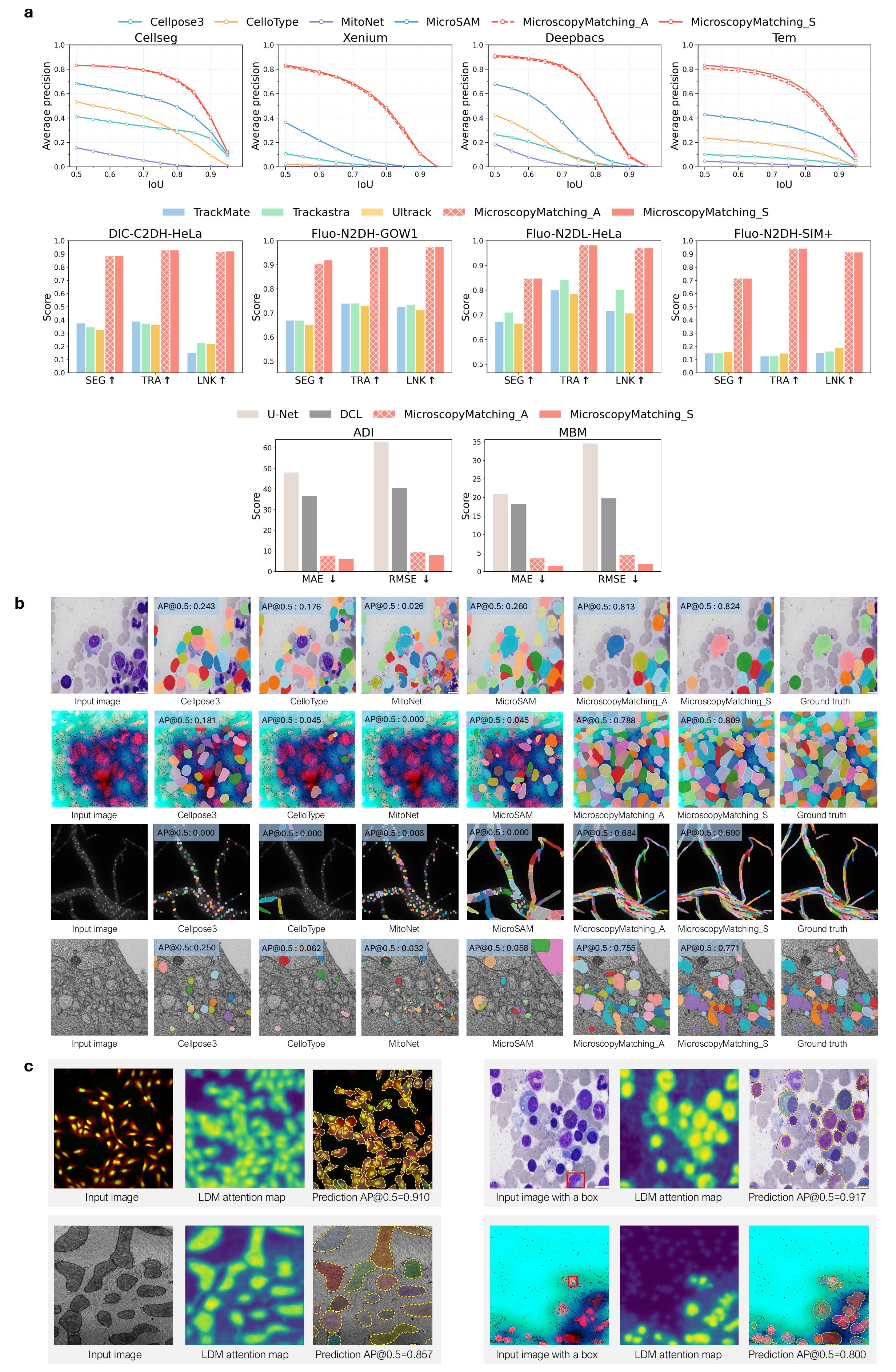}
\end{figure}
\FloatBarrier
\afterpage{
\captionof{figure}{
\textbf{Evaluation of the generalization performance of MicroscopyMatching on unseen benchmark datasets.} \textbf{(a)}, Quantitative results of MicroscopyMatching\_S and MicroscopyMatching\_A on the unseen benchmark datasets. The results follow the same visualization format as Fig.~\ref{fig:result1}b. We directly evaluate the performance using the trained framework from the framework formulation stage, without any further fine-tuning. As shown, MicroscopyMatching consistently achieved strong performance (e.g., with AP score at IoU threshold of 0.5 higher than 0.8 for segmentation, TRA higher than 0.9 for tracking, and MAE lower than 10 for counting), while all the compared methods displayed significantly reduced accuracy across datasets compared with the results in Fig.~\ref{fig:result1}b. \textbf{(b)}, Qualitative comparison of MicroscopyMatching and the compared methods on segmentation. For each example, we show the input image, the predictions of the compared methods and MicroscopyMatching, as well as the corresponding ground truth. For better visualization of the performance, the mask of each instance is displayed in a randomly assigned color. The image-level AP scores at an IoU threshold of 0.5 (i.e., AP@0.5) are shown on the top-left of each corresponding prediction. As shown, when applied to diverse and unseen microscopy images, the compared methods frequently broke down and failed to produce reliable results, while MicroscopyMatching consistently maintained reliable performance. More qualitative comparisons on tracking and counting are shown in Supplementary Figures 3--4.
\textbf{(c)}, Visualization results of the attention maps extracted from the LDM and MicroscopyMatching's corresponding predicted segmentation outputs on unseen datasets. Specifically, for each example, we show the input image, the attention map extracted from the LDM, and the predicted segmentations (in yellow dashed lines) together with the ground-truth segmentations (in purple lines) overlaid on the input image. For results obtained with MicroscopyMatching\_S mode, we also show the single bounding box (in red) containing the exemplar of the target object in the input image. The image-level AP score at an IoU threshold of 0.5 is shown below each visualized prediction. As shown, the LDM attention maps, serving as the matching results, can highlight the regions corresponding to the target objects, facilitating MicroscopyMatching to achieve good performance in unseen microscopy image analysis conditions. More qualitative results on segmentation, tracking, and counting tasks are shown in Supplementary Figure 5.
}
\label{fig:results2}}

In this evaluation, we directly compared MicroscopyMatching (without any additional fine-tuning) to the same methods described above. The performance comparisons are presented in Fig.~\ref{fig:results2}a and Supplementary Tables 4–6. As shown, without dataset-specific fine-tuning, MicroscopyMatching still achieved strong performance consistently on all tasks and benchmarks, and significantly outperformed all existing methods. Notably, the results also show that the compared methods exhibited a clear performance drop on unseen settings compared to their results on seen settings (Fig.~\ref{fig:result1}b), indicating that their performance is sensitive to the differences in the testing data from training data, e.g., different biological targets, different microscopy equipment, and new combinations of experimental conditions. To further illustrate the performance difference, we visualized the qualitative results between MicroscopyMatching and the compared methods in Fig.~\ref{fig:results2}b and Supplementary Figures 3--4. MicroscopyMatching consistently produced reliable results under different unseen experimental settings, while the compared methods performed poorly or even completely failed to produce meaningful results. This highlights MicroscopyMatching's practical advantage as a ready-to-use solution that requires no additional training or model adaptation which can be troublesome or even impractical for biomedical researchers.

We hypothesize that MicroscopyMatching's broad applicability stems from our reformulation of microscopy image analysis tasks as matching problems. To investigate this, we visualized the matching results (the attention maps) derived from the LDM in Fig.~\ref{fig:results2}c. As can be observed, across diverse testing samples with vastly different imaging conditions, the attention maps robustly identify image regions that correspond to the target objects. This robustness can be attributed to the nature of matching: since matching operates within individual images, it only requires establishing correspondence under the same imaging condition, rather than learning to recognize objects across diverse experimental settings. Consequently, based on this robust matching-based formulation, MicroscopyMatching can reliably operate across diverse experimental settings by performing within-domain matching in each image, while existing high-level understanding methods fail when conditions change. We show more qualitative results for segmentation, tracking, and counting in Supplementary Figure 5.

\subsection*{Evaluation on Collection of Image Data from Diverse, Real-world Experiments}

Beyond evaluation on standard benchmark datasets, to better reflect the real-world diversity of microscopy image analysis settings that biomedical researchers face in practice, we further assessed MicroscopyMatching's robustness under such real-world heterogeneity. To this end, we curated a 

\FloatBarrier
\begin{figure}[H]
    \centering
    \includegraphics[height=\textheight]{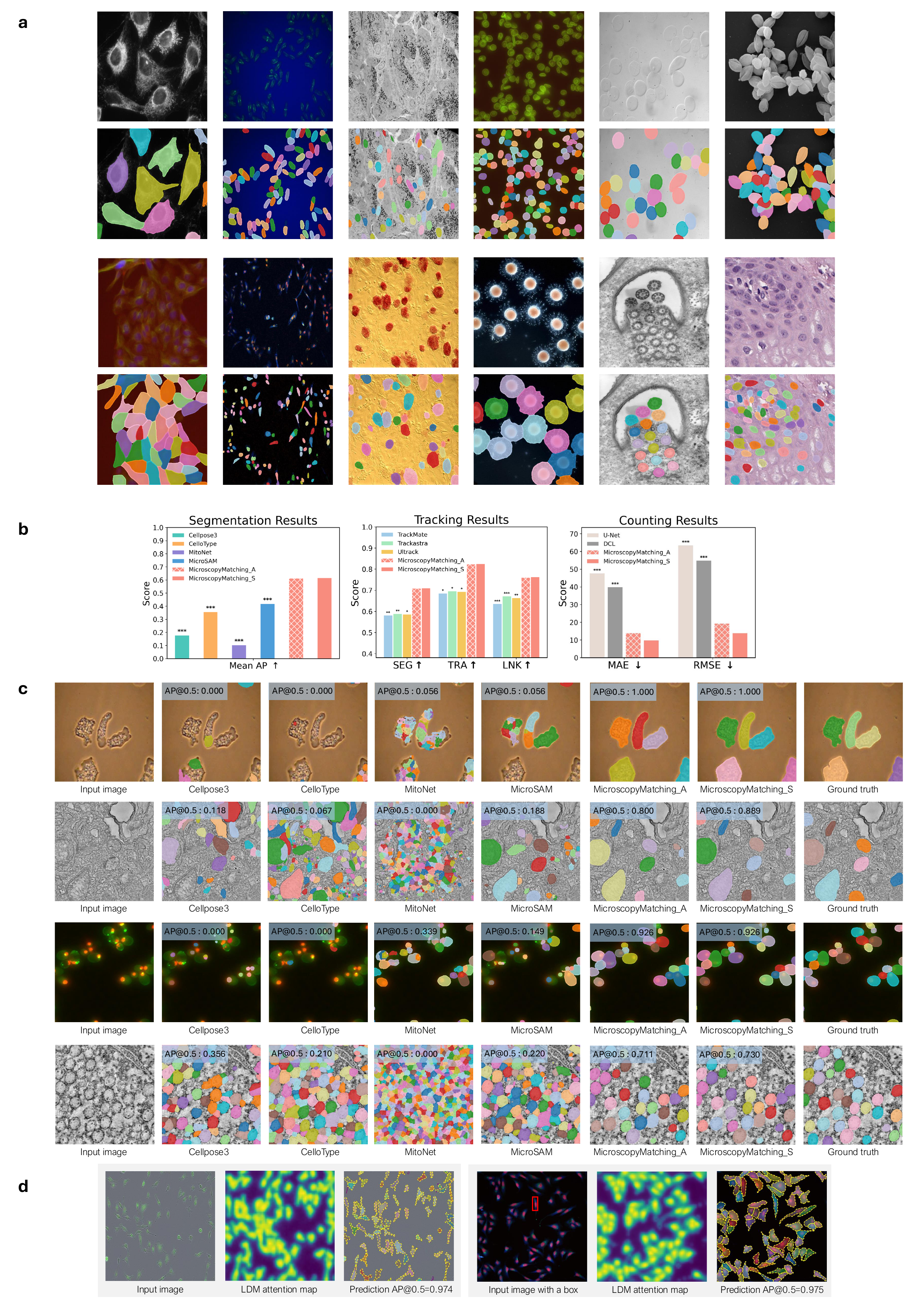}
\end{figure}
\FloatBarrier
\afterpage{
\captionof{figure}{
\textbf{Evaluation of MicroscopyMatching on the collection of image data from 200 sets of diverse, real-world experiments.} \textbf{(a)}, Examples of images from the image collection, shown with MicroscopyMatching\_A segmentation overlays. For better visualization, each predicted instance mask overlay is assigned a random color. \textbf{(b)}, Quantitative comparisons of MicroscopyMatching across segmentation, tracking, and counting tasks. For segmentation results, the bar plot shows the mean AP scores of each method (higher AP scores indicate better segmentation performance). For tracking results, the bar plot shows SEG, TRA, and LNK scores (higher scores indicate better tracking performance). For counting results, the bar plot shows MAE and RMSE scores (lower MAE and RMSE indicate better counting performance). Statistical significance of the performance improvement of MicroscopyMatching (in MicroscopyMatching\_A mode) over existing methods is indicated by asterisks above the corresponding bars of each compared method (* $P < 0.05$, ** $P < 0.01$, *** $P < 0.001$). P-values were computed using one-sided Student's t-test following \cite{pang2025cellotype}. See Methods for details, and see Supplementary Tables 7–9 for statistical significance results of the MicroscopyMatching\_S mode. \textbf{(c)}, Qualitative comparisons of MicroscopyMatching and other methods. For each example, we show the input image, the predictions of the compared methods and MicroscopyMatching, as well as the corresponding ground truth. The mask of each predicted instance is displayed in a randomly assigned color. The image-level AP scores at an IoU threshold of 0.5 (AP@0.5) are shown on the top-left of each corresponding prediction. As can be observed, the compared methods failed drastically under diverse experimental settings, whereas MicroscopyMatching remained robust. \textbf{(d)}, Visualization results of the attention maps extracted from the LDM and MicroscopyMatching's corresponding predicted segmentation outputs. Specifically, for each example, we show the input image, the extracted attention map, and the predicted segmentation (in yellow dashed lines) together with the ground-truth segmentation (in purple lines) overlaid on the input image. The image-level AP score at an IoU threshold of 0.5 is shown below each example.  For MicroscopyMatching\_S example, we also show the input exemplar bounding box (in red) containing the exemplar of the target object in the input image. More qualitative examples on tracking and counting are presented in Supplementary Figure 6.}
\label{fig:results3}
\vspace{4mm}
}

\noindent diverse, real-world collection of microscopy image data for evaluation, with each example drawn from a different microscopy analysis setting, varying in subject species, biological objects, sample preparation protocols, microscopy platforms, analysis goals, research facilities, etc., thereby capturing the heterogeneity of practical microscopy experimental scenarios (as shown in Fig.~\ref{fig:results3}a; see Methods for more details of the collection). In total, this collection comprises image data from 200 sets of distinct experiments, providing a realistic testbed for assessing MicroscopyMatching's robustness under diverse conditions. For this evaluation, we compared MicroscopyMatching\_A and MicroscopyMatching\_S with the same existing methods described above.

As shown in Fig.~\ref{fig:results3}b,c (and Supplementary Tables 7–9), under such diverse conditions, all the compared methods often failed drastically, whereas MicroscopyMatching consistently produced accurate predictions across this highly heterogeneous collection of image data on all tasks. This robust performance is further supported by the robust matching results visualized by the attention maps (Fig. \ref{fig:results3}d), which demonstrate that the underlying matching remains reliable across diverse and previously unseen conditions. More qualitative results of tracking and counting are presented in Supplementary Figure 6. Together, these results show that, in contrast to the compared methods that frequently broke down and failed to operate reliably in practice, MicroscopyMatching remained robust across highly heterogeneous microscopy experimental settings, consistently producing reliable outputs and supporting ready-to-use analysis.

\subsection*{MicroscopyMatching Achieves Human-level Performance}

Having demonstrated across extensive benchmarks (Fig. \ref{fig:result1}b and Fig. \ref{fig:results2}a) and diverse experimental conditions (Fig. \ref{fig:results3}b) that MicroscopyMatching consistently delivers strong and stable performance, we next ask whether these metric-level results translate into practical utility, in other words, whether MicroscopyMatching can actually help reduce the need for manual analysis (i.e., annotations by human annotators). A critical standard for this is whether MicroscopyMatching's outputs achieve a quality comparable to human annotations. We assess this with the following evaluations: expert rating and human-model agreement analysis.

\begin{figure}[H]
    \centering
    \includegraphics[width=\linewidth]{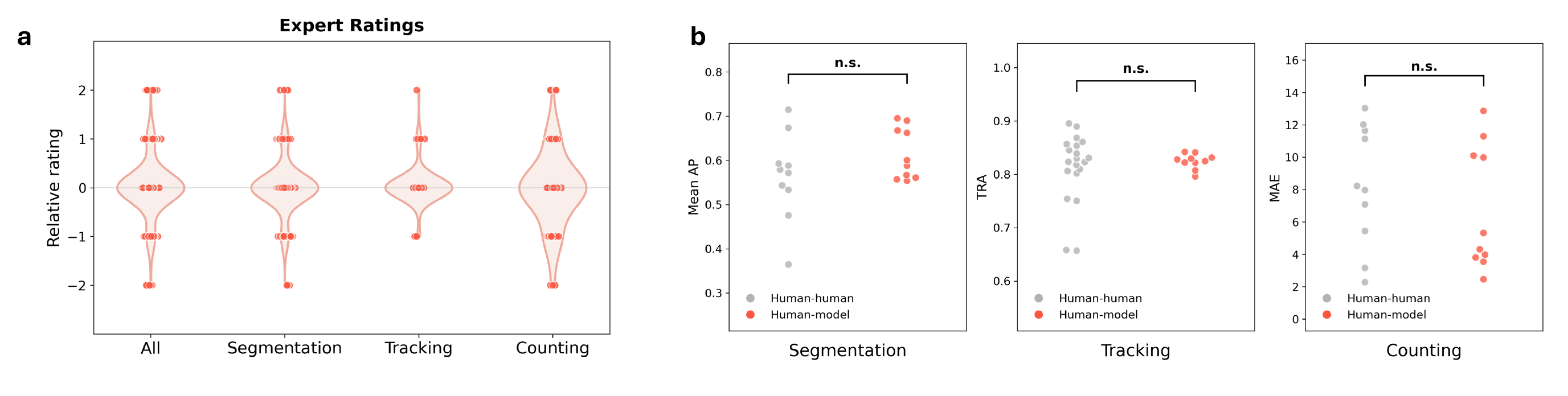}
    \caption{
    \textbf{Evaluation of MicroscopyMatching’s reliability relative to human annotators.} \textbf{(a)}, Expert ratings from the blinded evaluation comparing MicroscopyMatching's predictions and human annotations, where positive values in the rating indicate preference for MicroscopyMatching and negative values indicate preference for human annotations. In this evaluation, experts showed no significant preference toward human annotations or MicroscopyMatching's outputs ($P=0.44$; following \cite{greenwald2022whole}, P-value was computed using one-sample t-test), indicating that MicroscopyMatching achieves human-level performance. \textbf{(b)}, Human-model agreement evaluation of MicroscopyMatching following the protocol in \cite{greenwald2022whole}. Specifically, results from five independent annotators were compared both with one another (to measure human-human agreement) and with MicroscopyMatching’s predictions (to measure human-model agreement). Mean AP, TRA, and MAE are reported for segmentation, tracking, and counting tasks, respectively.     Following \cite{greenwald2022whole}, we conducted two-sample t-tests comparing human-human agreement and human-model agreement, and detected no significant difference (n.s. denoting not statistically significant; Supplementary Table 10), again demonstrating that MicroscopyMatching achieves human-level performance.
    }
    \label{fig:human_model}
    \vspace{3mm}
\end{figure}

For the first evaluation, we invited eight biomedical experts to conduct a blinded performance evaluation, comparing MicroscopyMatching's outputs with human annotations (more details in Methods). Each expert was shown the paired results (i.e., MicroscopyMatching's output and the corresponding human annotation) presented in random order, and was asked which result provided more accurate analysis. As shown in Fig. \ref{fig:human_model}a and Supplementary Table 10, the experts rated the two results comparably, with no statistically significant preference for human annotations or MicroscopyMatching's predictions, indicating that MicroscopyMatching performs at a level comparable to human annotators.

Besides expert ratings, to more meticulously evaluate MicroscopyMatching, we further conducted human-model agreement evaluations on the collection of diverse, real-world microscopy image data, where the samples span variations of diverse experiment conditions, closely reflecting real-world analysis scenarios. Following established human-model agreement evaluation protocol \cite{greenwald2022whole}, among the annotations produced by five independent human annotators, we assessed human-human agreement across all pairs of annotations by treating one annotator’s annotations as the ground truth and the other’s as the prediction. We then evaluated human-model agreement between MicroscopyMatching and each annotator by treating each of the five annotators’ annotations as the ground truth to assess MicroscopyMatching’s predictions. As shown in Fig.~\ref{fig:human_model}b, the human-model agreements fell within the variability range of human-human agreements. Following \cite{greenwald2022whole}, we further conducted two-sample t-tests comparing human–human agreement with human–model agreement, and detected no significant differences (Supplementary Table~10). Collectively, these results again demonstrate that MicroscopyMatching achieves human-level performance.

\subsection*{User Study of MicroscopyMatching}

\begin{figure}[H]
    \centering
    \includegraphics[width=0.9\linewidth]{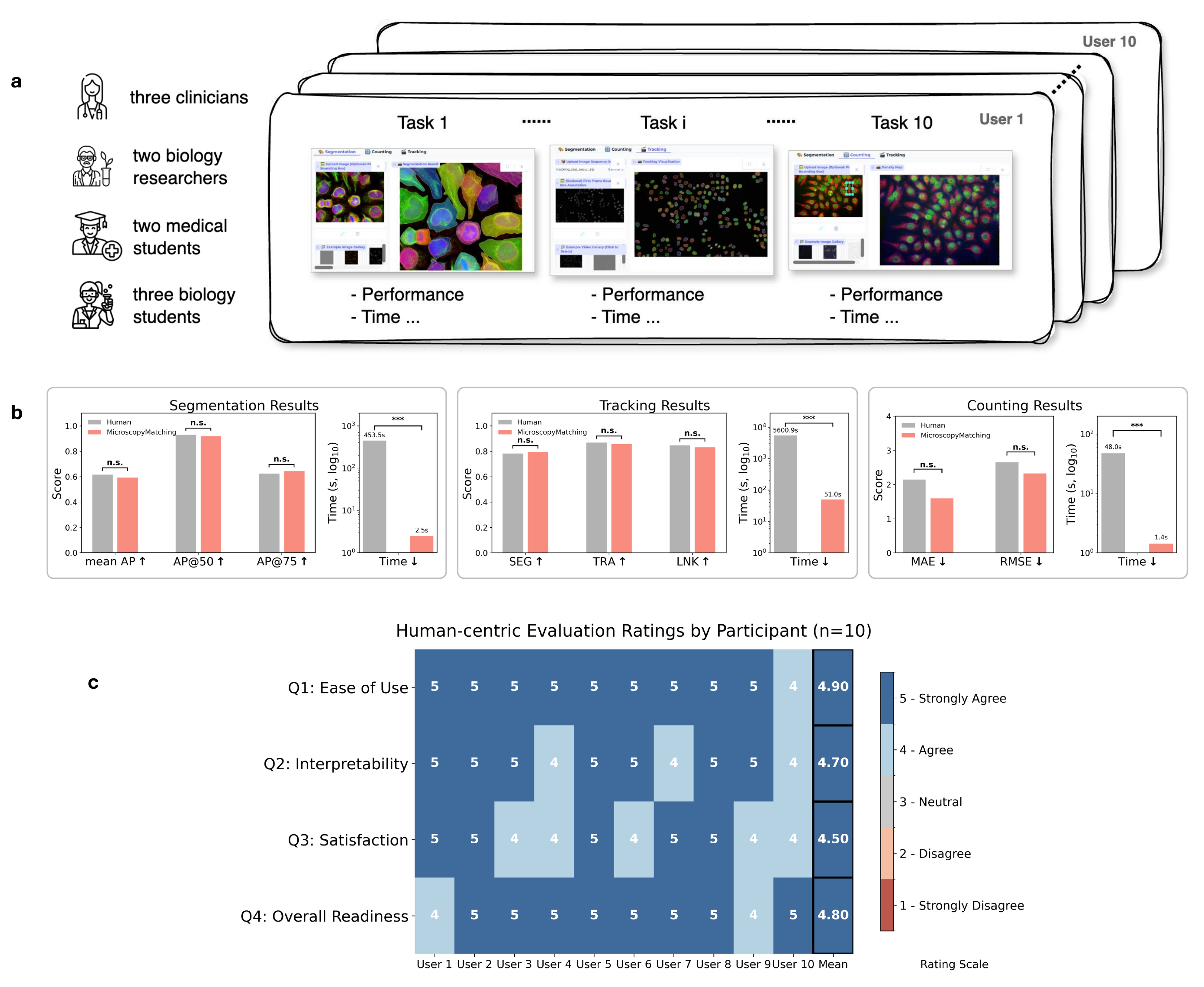}
    \caption{\textbf{Evaluating MicroscopyMatching by conducting a user study involving biomedical professionals with different backgrounds.} \textbf{(a)}, The user study involved ten biomedical professionals, including two biology researchers, three clinicians, two medical students, and three biology students. Each of these participants performed 10 microscopy analysis tasks using MicroscopyMatching, resulting in 100 test cases. The time to complete each task and the corresponding task performance were recorded. \textbf{(b)}, Comparison between the MicroscopyMatching tool and manual annotation in terms of task performance and completion time. P-values were computed using two-sided t-test \torevise{(Supplementary Table 11)}, n.s. indicates no significant differences between the two compared results, and asterisks indicate statistical significance level (*** $P < 0.001$).     Together, these results show that, in the user study, MicroscopyMatching significantly reduced task completion time, while achieving performance comparable to human annotation. \textbf{(c)}, Summary of user ratings for MicroscopyMatching in the post-task questionnaire. Participants rated the tool's ease of use, interpretability of the predictions, satisfaction with the performance, and their perceived overall readiness-of-use on a five-point Likert scale (1: strongly disagree, 2: disagree, 3: neutral, 4: agree, 5: strongly agree). The rightmost column shows the mean score for each question, with all aspects receiving high ratings, indicating a highly positive user experience for MicroscopyMatching.
    }
    \label{fig:user_study}
    \vspace{3mm}
\end{figure}

As MicroscopyMatching is designed to support biomedical researchers in real workflows, we also examined its usability and practical readiness through a user study involving ten biomedical professionals from diverse domains and technical backgrounds (more details in Methods). The participants included two biology researchers, three clinicians, two medical students, and three biology students, based across multiple countries and regions. Their expertise spanned ecological studies, clinical medicine, hospital laboratory diagnostics, and other areas. During the user study, each participant analyzed ten microscopy image samples representative of their routine research data using MicroscopyMatching. These use cases covered diverse microscopy image analysis settings, involving a wide range of research subjects (such as plants, avian species, microorganisms, and clinical specimens such as bone marrow, brain tumors, and blood smear samples), spanning diverse imaging modalities (such as brightfield, fluorescence, confocal, electron, and phase-contrast microscopy). The user study recorded the time each participant spent completing each task using our tool and the corresponding model performance, totaling $10\times 10=100$ test cases across all participants (Fig. \ref{fig:user_study}a). In addition, after completing the tasks, each participant filled out a post-task questionnaire evaluating their user experience, including ease of use, interpretability of the predictions, satisfaction with performance, and the overall readiness of our tool for practical use. The entire study took around 30 minutes on average per participant.

For reference, we compared MicroscopyMatching's task completion time and performance with human annotators to assess whether MicroscopyMatching can serve as a robust, ready-to-use alternative to manual annotation. Across all participants and tasks, on average, MicroscopyMatching led to a 99.1\% reduction in task completion time compared with human annotation, indicating a substantially reduced burden on researchers compared with manual annotation (Fig.~\ref{fig:user_study}b and Supplementary Table 11). Meanwhile, MicroscopyMatching also consistently achieved robust performance, with accuracy showing no significant difference from human annotation (see Fig. \ref{fig:user_study}b and Supplementary Table 11). These results demonstrate that MicroscopyMatching can substantially reduce the burden on biomedical researchers in microscopy analysis workflows, while achieving accuracy comparable to that of human annotators.

Beyond its strong performance, in the post-task questionnaires, participants also reported highly positive experiences of using MicroscopyMatching. As shown in Fig.~\ref{fig:user_study}c, all participants agreed or strongly agreed that MicroscopyMatching is easy to use and that its prediction results are easy to interpret. In addition, all participants agreed or strongly agreed that MicroscopyMatching provides satisfactory results. Overall, all participants reported high scores of 4 or 5 for the readiness rating reflecting how well they felt the tool could be practically used in their own workflows, indicating their positive perceptions of MicroscopyMatching’s practical utility.

Taken together, these results demonstrate that MicroscopyMatching provides a ready-to-use and easy-to-use tool for biomedical researchers and practitioners with diverse backgrounds, enabling reliable analysis across different real-world laboratory workflows and usage scenarios.

\subsection*{Practical Guidelines for Providing Exemplars}

Complementary to the above evaluations, which demonstrate MicroscopyMatching's practical usability, we further systematically examined how it performs under different practical usage patterns to inform real-world usage. Specifically, our framework can optionally take user-provided exemplar bounding boxes as input to guide the matching and facilitate performance. In the previous evaluations 

\FloatBarrier
\begin{figure}[ht]
    \centering
    \includegraphics[width=0.9\linewidth]{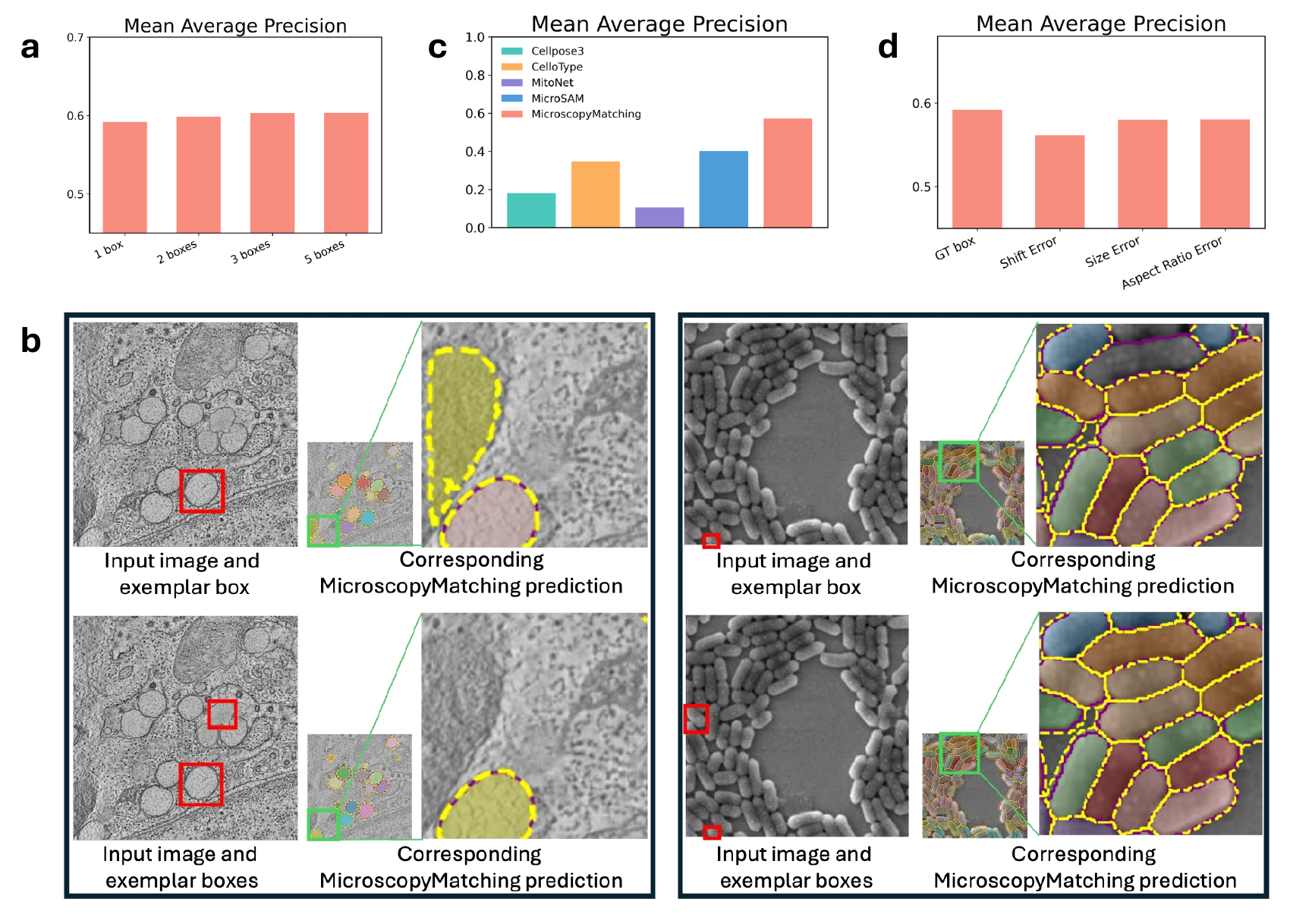}
    \caption{
    \textbf{Investigation of the practical usage of MicroscopyMatching.} \textbf{(a)}, Impact of providing different numbers of exemplar bounding boxes when using MicroscopyMatching. The bar plot shows performance when MicroscopyMatching is provided with 1, 2, 3, or 5 exemplar bounding boxes per test sample. \textbf{(b)}, Examples illustrating that providing additional exemplar boxes capturing variations of the target objects (e.g., in size or shape) can slightly enhance MicroscopyMatching's performance. For each example, input image with exemplar bounding boxes (in red), as well as the corresponding predictions (in yellow dashed lines) and the ground-truth segmentation (in purple lines) overlaid on the input image. The green boxes highlight the representative regions where the use of more exemplars brings improvements. Zoomed-in views of highlighted regions are shown on the right for clearer comparison. \textbf{(c)}, Bar plot showing the mean AP segmentation performance of the compared methods and MicroscopyMatching (with imprecise exemplar bounding boxes). Notably, even with imprecise bounding boxes, MicroscopyMatching still consistently outperformed all the compared methods significantly. \textbf{(d)}, We further report MicroscopyMatching’s performance (mean AP) under different types of exemplar bounding box imprecision. As shown, among the error types, shift error caused the greatest performance degradation.
    }
    \vspace{3mm}
\label{fig:practical_guideline}
\end{figure}

\noindent in Fig. \ref{fig:result1}–\ref{fig:results3}, we assessed MicroscopyMatching\_S where a single bounding box was provided for each testing sample, and the box was typically carefully specified by professional biomedical experts (i.e., ``ground truth'' bounding box). We now extend our analysis to systematically explore less controlled practical usage scenarios. Specifically, we investigate two explorations: (1) the impact of providing more than one bounding box per testing sample during inference, and (2) the impact of using bounding boxes that are less carefully specified and may contain various forms of imprecision.

In the first exploration, we investigated MicroscopyMatching’s performance with different numbers of exemplar bounding boxes (see Methods and Supplementary Note~1 for details). As shown in Fig.~\ref{fig:practical_guideline}a and Supplementary Table~12, providing additional exemplar boxes leads to modest performance gains. Moreover, we observe that when the additionally provided exemplar boxes capture variations of the target objects (e.g., size or shape differences), the performance improvement can be relatively more pronounced. In Fig.~\ref{fig:practical_guideline}b, we demonstrate the performance improvement when MicroscopyMatching is provided with additional exemplar boxes that capture such variations. Based on these observations, we recommend that, when seeking improved performance beyond a single exemplar box, users can consider adding more bounding boxes, particularly those capturing variations of the target objects.

In the second exploration, considering that imprecision of the exemplar bounding box may happen in practical usage, we investigated the impact of using exemplar bounding boxes that were less carefully specified (more details about this investigation are in Methods). The results (Fig.~\ref{fig:practical_guideline}c and Supplementary Table 12) show that MicroscopyMatching still outperformed the compared methods significantly even when exemplar bounding boxes were less precisely specified. To characterize MicroscopyMatching's sensitivity to exemplar box specification imprecision, we further examined MicroscopyMatching's robustness to different types of imprecision that users may introduce during specifying the exemplar bounding box. Following \cite{bolya2020tide,strafforello2022humans,zheng2020distance}, bounding box imprecision was categorized into three types: shift error, size error, and aspect ratio error (Methods). We then evaluated MicroscopyMatching's performance under each error type (Fig.~\ref{fig:practical_guideline}d and Supplementary Table 12). As shown, among the imprecision types, shift error resulted in the greatest performance degradation. From this, we recommend users to exercise particular care to reduce shift-error mistakes when using MicroscopyMatching.

Overall, these explorations provide practical guidelines to help users apply MicroscopyMatching more effectively in real-world scenarios.

\section*{Discussion}\label{sec:discussion}

In the present study, we introduce MicroscopyMatching, a ready-to-use framework for microscopy image analysis in practice. In contrast to existing microscopy image analysis methods that frequently fail under the highly diverse and ever-evolving microscopy image analysis settings, we show that MicroscopyMatching performs reliably across diverse and previously unseen experimental settings, without requiring burdensome efforts of model adaptation for each specific microscopy experimental scenario. First, across 20 different benchmark datasets spanning a wide variety of microscopy image analysis settings, MicroscopyMatching consistently delivers superior performance. Second, beyond standard benchmarks, MicroscopyMatching remains robust on an even more heterogeneous collection of microscopy image data from diverse real-world experiments, consistently delivering reliable performance under practical experimental variability. Third, to investigate whether this performance reaches a practically meaningful level, we conducted expert rating and human-model agreement evaluation, which further demonstrate that MicroscopyMatching achieves performance comparable to human annotators. Finally, a user study shows that MicroscopyMatching is practically useful and helpful for biomedical researchers across diverse backgrounds, further supporting MicroscopyMatching as a ready-to-use tool for microscopy image analysis. Taken together, results across these complementary evaluation dimensions establish MicroscopyMatching as a reliable and ready-to-use framework for microscopy image analysis in practice.

Despite the demonstrated practical effectiveness of our MicroscopyMatching tool, there are several limitations that suggest directions for future work. First, in this work, we focus on microscopy image analysis, a domain of critical importance. However, it would also be valuable to investigate whether the underlying insights of MicroscopyMatching can be extended to other biological imaging modalities, such as X-ray, magnetic resonance, and ultrasound imaging. Second, regarding the choice of pre-trained LDM, we adopted Stable Diffusion \cite{rombach2022high}, a widely used and powerful model, as the foundation of MicroscopyMatching. While this choice proved very effective, the rapid evolution of LDMs in the AI field makes it meaningful and worthwhile for follow-up studies to explore the integration of more advanced LDMs into the framework. Finally, MicroscopyMatching currently handles non-two-dimensional microscopy data (such as volumetric electron microscopy images) indirectly by first reducing them to two-dimensional representations, a strategy that is commonly adopted in biomedical analysis pipelines \cite{zhao2025foundation}. Extending MicroscopyMatching to directly handle three-dimensional data would also be valuable for future work.

\bibliography{main}

\section*{Methods}

\subsection*{MicroscopyMatching}
The overview of framework architecture of MicroscopyMatching is demonstrated in Fig.~\ref{fig:result1}a. From a high level, the framework consists of four parts: (1) a pre-trained variational autoencoder (VAE) to encode the input microscopy image to latent image features; (2) an embedding encoder module to obtain the conditioning embedding used in the matching process; (3) a pre-trained diffusion module that processes the latent image feature and the conditioning embedding and produces the attention maps; (4) the attention post-processing modules to predict the final results based on the attention maps. Our MicroscopyMatching framework is constructed based on Stable Diffusion \cite{rombach2022high}, a widely-used and powerful latent diffusion model (LDM) originally trained on the LAION-2B-en dataset \cite{schuhmann2022laion}. Within the overall architecture, the VAE and the diffusion module are adopted from the pre-trained LDM and their parameters are kept fixed. The exemplar projector in the embedding encoder (detailed below when describing the embedding encoder), together with the attention post-processing modules, are trainable. Below we first introduce these modules in details respectively, then describe the overall pipeline of the two modes of MicroscopyMatching, i.e., MicroscopyMatching\_S and MicroscopyMatching\_A.

\noindent\textbf{Pre-trained VAE.} Given an input microscopy image $x$, it is first sent into the pre-trained VAE $\mathcal{E}$ to produce the latent image feature $z=\mathcal{E}(x)$. The VAE module is built based on convolutional autoencoder architecture. We adopt the pre-trained VAE module from Stable Diffusion, which has been extensively trained on billions of diverse images (in LAION-2B-en) and has learned a meaningful latent space. During the training of MicroscopyMatching, the VAE parameters are kept frozen.

\noindent\textbf{Embedding encoder.} The embedding encoder generates the conditioning embedding in two modes: when provided with a bounding box containing an exemplar of the target objects (the MicroscopyMatching\_S mode), it projects the given exemplar to an exemplar embedding; if no exemplar bounding box is provided (MicroscopyMatching\_A mode), it generates an internal embedding to trigger automatic repetitive pattern identification and matching. Below we introduce these processes respectively.

In MicroscopyMatching\_S, an exemplar is specified by a bounding box $b_{\text{exe}}$ on the input image $x$, which contains an example of the target biological object. To encode this exemplar to obtain the conditioning embedding, we employ an exemplar projector $\phi_E$, which is built based on a convolutional neural network (CNN) architecture, with SwAV-pre-trained \cite{caron2020unsupervised}  ResNet-50 network \cite{he2016deep} as the backbone. Specifically, given the input image $x$ and exemplar box $b_{\text{exe}}$, the exemplar projector first extracts image feature $f_x$. It then obtains the exemplar feature $f_{\text{exe}}$ by applying region of interest (RoI) pooling \cite{girshick2015fast} corresponding to $b_{\text{exe}}$ on the image feature. Finally, the exemplar feature is processed by a $3 \times 3$ convolution layer and a linear layer to produce the conditioning embedding $e\in \mathbb{R}^{d_e}$, where $d_e$ represents the dimension of the embedding $e$ and $d_e$ is set to 768.

In MicroscopyMatching\_A, no bounding box exemplar is provided, and the framework operates in an automatic mode where a fixed template $y$ (``repetitive objects'') is used to trigger the semantic matching within the image  (see Supplementary Note 2 for more details about the template). The template is fed into the pre-trained text encoder $\tau_\theta$ of Stable Diffusion to produce the conditioning embedding $e = \tau_\theta(y)$, with $e \in \mathbb{R}^{d_e}$. We adopt the pre-trained Stable Diffusion text encoder and keep its parameters fixed; it follows the CLIP ViT/L architecture \cite{radford2021learning} and has an embedding dimension $d_e = 768$.

\noindent\textbf{Pre-trained diffusion module.}
Given the latent image feature $z$ (extracted by the VAE) and the conditioning embedding $e$ (obtained from the embedding encoder) as inputs, the pre-trained diffusion module $\varepsilon_\theta$ performs the conditional diffusion process, where the matching process is conducted and attention maps capturing the fine-grained matching results are extracted. We adopt the pre-trained diffusion module, i.e., the denoising U-Net, from Stable Diffusion, and keep its parameters fixed. Below, we describe the matching process in the MicroscopyMatching\_S mode and the MicroscopyMatching\_A mode, respectively.

\noindent\underline{Matching Process in MicroscopyMatching\_S.}
In MicroscopyMatching\_S, as an exemplar bounding box containing the target object is provided, the matching process is performed by finding correspondence between the exemplar and the entire input image. Specifically, this is achieved by leveraging the self- and cross-attention mechanisms in the pre-trained LDM. First, following the standard diffusion process \cite{rombach2022high}, we add standard Gaussian noise $\epsilon$ with respect to the diffusion time step $k$ to the latent image feature $z$ to obtain the noisy latent $z_k$. Then, the noisy latent $z_k$ is sent into the denoising U-Net, along with the exemplar embeddings $e$ as the condition, to perform the conditional denoising process. During this process, the interaction between the image content and the conditioning embeddings is modeled in the cross-attention layers of the denoising U-Net \cite{rombach2022high}. Notably, in its $l$-th cross-attention layer, the query $Q_c^l$ is computed from the noisy image latent $z_k$, while the key $K_c^l$ and value $V_c^l$ are respectively derived from the conditioning embeddings $e$. The output of this cross-attention layer is computed as $\text{SoftMax}(\frac{Q_c^l K_c^l}{\sqrt{d_l}})\cdot V_c^l$, where $d_l$ is the dimension of $Q_c^l$, $K_c^l$ and $V_c^l$. We then extract the cross-attention map from the $l$-th cross-attention layer as $M_c^l$:
\begin{equation}
    M_{c}^l = \text{SoftMax}(\frac{Q_c^l K_c^l}{\sqrt{d_l}}).
\end{equation}
By collecting $M_c^l$ across different cross-attention layers and averaging them, we can get the cross-attention map $M_c \in \mathbb{R}^{h\times w}$. 

Besides cross-attention maps, we also incorporate self-attention maps as a complementary signal to enhance within-object coherence. Specifically, in the $m$-th self-attention layer, the query $Q_s^m$, key $K_s^m$, and value $V_s^m$ are all derived from the image content, i.e., the noisy image latent $z_k$. Similar to cross-attention maps, the self-attention map in the $m$-th self-attention layer can be obtained as $M_s^l=\text{SoftMax}(\frac{Q_s^m K_s^m}{\sqrt{d_m}})$, where $d_m$ represents the dimension of $Q_s^m$, $K_s^m$, and $V_s^m$. We also collect and average $M_s^l$ across different layers to obtain the self-attention map $M_s \in \mathbb{R}^{h\times w \times h \times w}$. Because self-attention maps are computed by modeling the correlation between the query and key that are both derived from the image content, they can encode the semantic association between each pixel and other pixels in the image. Intuitively, pixels belonging to the same object type tend to have higher association. This within-image correlation information in self-attention maps can complement the semantic correspondence between the conditioning input and image regions captured by the cross-attention maps, facilitating more precise and robust matching. Specifically, we apply the cross-attention map $M_c$ to the self-attention map $M_s$ to compute the self-cross attention map $M_{sc}\in\mathbb{R}^{h\times w}$ via:
\begin{equation}
    M_{sc} = \sum_{p=1}^{h} \sum_{q=1}^{w} M_c(p,q)M_s(p,q,:,:),
    \label{eq:self_cross}
\end{equation}
where $M_c(p,q)$ is the cross-attention value of pixel location $(p,q)$, and $M_s(p,q,:,:)\in \mathbb{R}^{h\times w}$ is the 2D self-attention map corresponding to pixel location $(p,q)$. The cross attention map $M_c$ and cross-self attention map $M_{sc}$ are passed into the following attention post-processing module to predict the final results.

\noindent\underline{Matching Process in MicroscopyMatching\_A.}
In MicroscopyMatching\_A mode, the model performs matching directly within the input image, without requiring an exemplar bounding box, instead taking the fixed semantic template (``repetitive objects''). In the absence of an explicit exemplar box, the matching process relies on intrinsic within-image cues to discover and match the target objects. Specifically, MicroscopyMatching\_A leverages the rich within-image correlation information encoded in the self-attention map $M_s$. As elaborated above, the self-attention layers model pairwise relationships between each pixel and all other pixels in the input image latent, yielding a dense correlation map. Pixels belonging to the same object or the same object type, which share similar semantic properties, tend to exhibit stronger mutual associations in these maps. This essentially captures grouping information indicating which pixels likely belong to same underlying objects, providing informative cues for matching. Yet, while informative, it reflects grouping information across all image contents. To perform matching between target objects, we combine it with auxiliary cues. Specifically, to trigger the model to automatically discover and match visually recurring structures without using exemplar bounding boxes, we incorporate cross-attention map $M_c$ (obtained using the above derived internal embedding as the conditional embedding) to compute the self-cross attention map $M_{sc}$ as described in Eq. \ref{eq:self_cross}. This self-cross attention map $M_sc$ integrates pixel grouping information (from $M_s$) with pattern recurrence cues (from $M_c$), enabling the model to automatically identify and match recurring object instances. Consistent with the MicroscopyMatching\_S mode, both the cross-self attention map and the cross-attention map are passed to the shared attention post-processing module.

\noindent\textbf{Attention post-processing module.}
As different tasks (segmentation, tracking, and counting) require different outputs (segmentation masks, tracking trajectories, and object counts), we employ task-specific attention post-processing modules. These attention post-processing modules are shared across both the MicroscopyMatching\_S and MicroscopyMatching\_A usage modes. For segmentation, the attention post-processing module predicts the instance segmentation mask for each given input image, with each object instance delineated separately. The attention post-processing module consists of a lightweight processing stage followed by a Transformer-based encoder and a convolutional prediction head. We first take the attention maps and image features (extracted from the pre-trained diffusion module) and pass them through a normalization layer and a convolution layer to obtain a fused representation. A Vision Transformer (ViT) encoder \cite{li2022exploring}, configured with a $14\times 14$ attention window and four global attention layers following \cite{pachitariu2025cellpose}, is then applied to model global context while preserving spatial layout. The output tokens from the encoder are reshaped back into a spatial feature map and further refined by two convolution layers to produce the final segmentation mask.

For tracking, the attention post-processing module aims to predict both the segmentation masks of the objects in each frame and their linkage across time. Specifically, the attention post-processing module comprises the segmentation prediction submodule and the linkage prediction submodule. The segmentation prediction submodule has the same architecture as the segmentation attention post-processing module described above. The linkage prediction submodule first processes the attention maps through a normalization layer, a convolution layer, and a multilayer perceptron (MLP) layer to obtain compact representations. The representations are then fed into an encoder-decoder Transformer architecture along with low-dimensional feature vectors (encoding object properties such as position, size, and appearance) following \cite{gallusser2024trackastra}. The Transformer predicts an association probability matrix between detected objects in consecutive frames to establish inter-frame correspondences. The final outputs of the attention post-processing module for tracking include the segmentation masks for each frame, in which the same object appearing across different frames is assigned a consistent instance ID, together with an acyclic graph to represent the temporal associations among all detected objects. These outputs follow the Cell Tracking Challenge (CTC) format \cite{mavska2023cell}.

For counting, the attention post-processing module produces a density map for each input image, where the pixel values in the map represent local object density, and the integral over the density map corresponds to the predicted object count. This attention post-processing module is built based on a CNN architecture \cite{djukic2023low}, with three layers of $3\times 3$ convolution layers, each followed by a Leaky ReLU layer and a bilinear upsampling layer with factor 2, as well as a $1\times 1$ convolution layer followed by a Leaky ReLU layer.

\noindent\textbf{Overall pipeline.}
MicroscopyMatching supports both usage modes: users can choose to provide a bounding box to specify a target object in the image as an exemplar for semantic matching (i.e., the MicroscopyMatching\_S mode), or let the model automatically identify and match recurring objects within the image (i.e., the MicroscopyMatching\_A mode). Below we first describe the MicroscopyMatching\_S mode in segmentation, tracking, and counting, then introduce the MicroscopyMatching\_A mode accordingly.

Under the MicroscopyMatching\_S mode, user provides a single bounding box specifying an exemplar of a target object in the image (or in the first frame in the case of tracking), which derives the conditioning embedding for subsequent matching and prediction. Specifically, for segmentation, given a testing image, MicroscopyMatching first transforms the image into image latent via the pre-trained VAE. The bounding box and the image is sent into the exemplar projector to compute the conditioning embedding $e$. The produced image latent and conditioning embedding are passed into the pre-trained diffusion module, where we obtain the cross-attention map and the self-cross attention map. We also extract image features from the pre-trained diffusion module to provide more image information. The attention post-processing module then produces the segmentation mask as described above. For counting, the overall pipeline remains similar, where the attention post-processing module produces density maps. The predicted number of the target objects is calculated by summing the pixel values in the predicted density map.

Different from segmentation and counting, for tracking, the input to the framework is a microscopy image sequence $[I_1, I_2, ..., I_T]$ (where $T$ is the total number of image frames). The goal of the framework is to both identify the target objects in each frame based on the conditioning embedding, and associate the objects across different frames. This process is essentially a dual-matching problem, involving both matching within each frame to identify the target objects, and matching across consecutive frames to establish temporal correspondences. (1) \textit{Matching within each frame to identify and segment the target objects:} Specifically, given the first frame, using conditioning embedding (derived from the exemplar bounding box), the framework predicts the segmentation mask via the segmentation submodule described above and initializes object identities. Then, for each subsequent frame $I_t$ ($t>1$), as we have access to the $n_{t-1}$ segmented objects $\{o_1^{t-1}, o_2^{t-1}, ..., o_{n_{t-1}}^{t-1}\}$ in the previous frame $I_{t-1}$, we can leverage these segmented objects as exemplars to obtain segmentation mask for frame $I_t$. More specifically, we use the $n_{t-1}$ segmented objects $\{o_1^{t-1}, o_2^{t-1}, ..., o_{n_{t-1}}^{t-1}\}$ in $I_{t-1}$ to derive $n_{t-1}$ tight bounding boxes $\{b_1^{t-1}, b_2^{t-1}, ..., b_{n_{t-1}}^{t-1}\}$ containing the objects. The image $I_{t-1}$ and the bounding boxes are then sent into the exemplar projector described above, producing $n_{t-1}$ embeddings ${e_1^{t-1}, e_2^{t-1}, ..., e_{n_{t-1}}^{t-1}}$. To obtain a representative conditioning embedding for the target object, we average these embeddings to obtain the embedding $e$. Then, the image latent of the current frame $t$ and embedding $e$ are passed to the pre-trained diffusion module to obtain the attention maps, which are then processed by the segmentation submodule to produce segmentation mask of frame $t$. (2) \textit{Matching across consecutive frames to establish temporal correspondences:} On the other hand, to predict the objects' temporal linkage, we perform object-to-object matching between consecutive frames. Specifically, given the current frame $I_t$ and the predicted segmented objects $\{o_1^{t-1}, o_2^{t-1}, ..., o_{n_{t-1}}^{t-1}\}$ in its previous frame, we perform matching between each segmented object $o_i^{t-1}$ and the current frame $I_t$ to obtain attention maps. This process yields object-specific attention maps that capture the spatial correspondence between the exemplar from the previous frame and its potential matches in the current frame. For this matching, we use only cross-attention without combining with self-cross attention, as the goal is direct object-to-region correspondence rather than discovering recurring patterns within a single frame. To improve matching robustness and reduce direction-specific bias, we perform this matching in both directions, i.e., from the previous frame to the current frame and vice versa. These attention maps are then passed into the linkage prediction submodule described above to predict the association matrix $A_t$. After aggregating the predicted association matrices across the image sequence, we obtain the final predicted tracking graph following \cite{gallusser2024trackastra}. The overall results are presented in CTC format, consisting of a text file recording the object IDs, temporal correspondences, and lineage relationships, along with the instance segmentation masks for each frame.

In the MicroscopyMatching\_A mode, the framework leverages the internal embedding to trigger the matching process for segmentation, tracking, and counting tasks. For all the three tasks (segmentation, tracking, and counting), the remaining pipeline follows the same procedure as described for the MicroscopyMatching\_S mode.

\subsection*{Implementation of MicroscopyMatching}

The trainable modules of MicroscopyMatching include: the attention post-processing module for segmentation, tracking, and counting (shared for both MicroscopyMatching\_S and MicroscopyMatching\_A modes), and the exemplar projector (used only in the MicroscopyMatching\_S mode).

To train the attention post-processing module for segmentation, we adopt the segmentation datasets including Cellpose Cyto, TissueNet, LiveCell, and Lucchi as the training datasets. We use the binary cross-entropy loss computed between the predicted segmentation mask and the ground-truth mask to supervise the training process following \cite{stringer2025cellpose3,pachitariu2025cellpose}. We adopt the AdamW optimizer with a learning rate of $1e-4$ and the cosine annealing scheduler. To train the attention post-processing module for tracking, we use the tracking datasets, including BF-C2DL-MuSC, BF-C2DL-HSC, Fluo-C2DL-MSC, and PhC-C2DH-U373 as the training datasets. The loss for the segmentation submodule is computed as cross-entropy between the predicted segmentation mask and the ground-truth mask, and the loss for the linkage prediction submodule is computed as the binary cross-entropy loss between the predicted association matrix and the ground-truth association matrix (derived from the ground-truth track graph following \cite{gallusser2024trackastra}). We adopt the AdamW optimizer with a learning rate of $1e-4$ and the cosine annealing scheduler. To train the attention post-processing module for counting, we use the counting datasets, including DCC and VGG. We compute the mean square loss between the predicted density map and the ground-truth density map. The ground-truth density maps are generated via convolving ground-truth dot maps (where each dot represents the center location of an object) with a Gaussian kernel following \cite{zheng2024rethinking}. We adopt the AdamW optimizer with learning rate of $1e-4$ and the cosine annealing scheduler. Across the training process for each of the three tasks, the model is evaluated on the validation set at the end of every training epoch. Training is terminated when the evaluation performance does not improve for 15 consecutive epochs, following \cite{pang2025cellotype}.

To train the exemplar projector for the MicroscopyMatching\_S mode, we adopt the segmentation datasets including Cellpose Cyto, TissueNet, LiveCell, and Lucchi as the training datasets. Specifically, during this training process, given a training image and its bounding box containing the image exemplar, we obtain the conditioning embedding $e$ via the exemplar projector, which is then used as the condition to obtain the cross-attention map $M_c$. Ideally, the cross-attention map should indicate the presence regions of the target objects, which can be derived from the ground-truth segmentation mask. Specifically, for supervision, the ground-truth segmentation masks are converted into binary maps (object = 1, background = 0), and the cross-entropy loss is computed by comparing the normalized cross-attention maps with the normalized versions of these binary maps, encouraging attention to focus on the true object regions. We adopt the AdamW optimizer with learning rate of $1e-6$ and the cosine annealing scheduler, and train the module for 200 epochs. 

Our experiments are conducted on NVIDIA RTX 6000 Ada GPUs. In our experiments, we adopt Stable Diffusion, in specific, its SD v1.4 version (\url{https://huggingface.co/CompVis/stable-diffusion-v1-4}), as the pre-trained LDM in MicroscopyMatching. We provide experiments investigating incorporating other Stable Diffusion versions in Supplementary Note 3. To obtain the noisy image latent $z_k$, we apply the noise to the image latent $z$ using the DDPM scheduler \cite{ho2020denoising} at $k=20$. The input image to MicroscopyMatching is resized to $512 \times 512$. For the MicroscopyMatching\_S usage mode, besides the input image, the framework also takes an exemplar bounding box as the input. For the experiments on the benchmark datasets, we obtain the exemplar bounding boxes in the following manner. For segmentation datasets, the exemplar bounding box is derived from a randomly selected annotated instance mask in each image. For tracking datasets, the exemplar bounding box is obtained from a randomly selected instance mask in the first frame of each image sequence. For counting datasets, the exemplar bounding box is constructed by randomly selecting an object instance and manually annotating the corresponding region. For experiments on the collection of images from diverse experiments, exemplar bounding boxes are constructed by randomly selecting object instances and manually annotating. The MicroscopyMatching framework is implemented in Python 3, using packages including pytorch, numpy, scipy, tifffile, skimage, and tqdm. The tool interface additionally uses gradio and gradio\_bbox\_annotator. The result figures were made using Matplotlib.

\subsection*{Metrics}
\noindent\textbf{Segmentation.} 
The average precision (AP) metric is a standard and widely adopted metric to evaluate the performance of instance segmentation in microscopy images \cite{stringer2021cellpose,pachitariu2022cellpose,stringer2025cellpose3,pang2025cellotype}. Specifically, the AP metric is computed by matching the predicted mask of each instance to the ground-truth instance mask that has the highest intersection over union (IoU) with the predicted instance mask. Given an IoU threshold (e.g., 0.50), a predicted mask for an instance is considered a true positive (TP) if its IoU with the corresponding ground-truth mask is greater than the IoU threshold; otherwise, it is considered a false positive (FP). The ground-truth masks that have no matched predicted masks are considered as false negatives (FN). AP is computed as $AP = \frac{TP}{TP+FP+FN}$. A higher AP score indicates better segmentation performance. For a comprehensive evaluation, the AP is calculated at varying thresholds of IoU (ranging from 0.50 to 0.95 with a step size of 0.05). We follow the implementation of AP metric in \cite{stringer2021cellpose,pachitariu2022cellpose,stringer2025cellpose3} to report AP for all the segmentation methods.

\noindent\textbf{Tracking.}
We evaluate the tracking performance using the tracking metrics in the well-established Cell Tracking Challenge \cite{mavska2023cell}, i.e., the tracking accuracy measure (TRA), the linking accuracy measure (LNK), and the segmentation accuracy measure (SEG). Specifically, TRA quantifies the overall tracking performance by jointly evaluating object detection and temporal association. It is computed by comparing the directed acyclic graph of the predicted object trajectories with the ground-truth tracking graph, penalizing errors such as false positives, missed detections, and incorrect temporal links. The higher TRA score suggests better overall tracking performance. LNK evaluates the temporal association performance by focusing solely on the correctness of object linkages between consecutive frames. It compares the predicted and ground-truth trajectory graphs in terms of their linkage edge, and reflects how accurately objects are followed over time. The higher LNK score indicates the better performance of correctly associating objects over time. SEG, on the other hand, assesses the segmentation quality by comparing the predicted and ground-truth object masks based on Jaccard similarity index, with a higher SEG score reflecting better performance of localizing the objects in each frame. Together, these metrics comprehensively evaluate detection accuracy, temporal association correctness, and overall tracking integrity.

\noindent\textbf{Counting.}
We use the widely-adopted metrics, mean absolute error (MAE) and root mean square error (RMSE) \cite{sam2017switching,djukic2023low,hui2024class} to evaluate the counting performance. These metrics measure the discrepancy between the predicted object counts and the ground-truth object counts. Specifically, MAE is calculated as $\text{MAE}=\frac{1}{N} \sum_{i=1}^N |c_{\text{pred}}^i-c_{\text{gt}}^i|$, where $c_{\text{pred}}^i$ and $c_{\text{gt}}^i$ denote the predicted count and the ground-truth count of the $i$-th sample, respectively, and $N$ is the number of tested samples. RMSE is calculated as $\text{RMSE}=\frac{1}{N} \sqrt{\sum_{i=1}^N (c_{\text{pred}}^i-c_{\text{gt}}^i)^2}$. A lower MAE or RMSE score indicates better counting performance, with MAE emphasizing average error and RMSE emphasizing larger errors.

\subsection*{Datasets}
\noindent\textbf{Segmentation datasets}

\noindent\textit{Cellpose Cyto} \cite{stringer2021cellpose} consists of images from various sources, such as fluorescence images of cultured neurons with both cytoplasmic and nuclear stains, fluorescence images of cells with cytoplasmic markers, brightfield microscopy images of nonfluorescence cells, fluorescence images of cells with membrane maker. This dataset is used both for training and evaluation.

\noindent\textit{TissueNet} \cite{greenwald2022whole} consists of multiplex microscopy images of tissue samples processed under multiple immunofluorescence staining protocols via six imaging platforms, i.e., CODEX, CyCIF, Vectra, MIBI-TOF, MxIF, and IMC. We use the cellular segmentation for each image. This dataset is used both for training and evaluation.

\noindent\textit{LiveCell} \cite{edlund2021livecell} contains phase-contrast microscopy images of eight cultured cell lines, acquired without fluorescent labeling. Each image is provided with instance-level segmentation annotations delineating individual cells. This dataset is used both for training and evaluation.

\noindent\textit{Lucchi} \cite{lucchi2012structured} contains focused ion beam scanning electron microscopy (FIB-SEM) image stacks of neural tissue, annotated at the slice level for individual mitochondria. Each 2D image slice is provided with instance-level segmentation annotations of individual mitochondria. This dataset is used both for training and evaluation.

\noindent\textit{NeurIPS CellSeg} \cite{ma2024multimodality} consists of microscopy images obtained from diverse imaging platforms and tissue types. It includes microscopy images that are diverse along four dimensions: cell origins (e.g., diverse tissue samples and cell lines cultured under different conditions), staining protocols (e.g., cells with various stains and label-free cells), microscope types (e.g., brightfield, fluorescent, phase-contrast, and differential interference contrast) and cell morphologies (e.g., round cells, elongated cells, and cells with irregular shapes). We adopt the test set of NeurIPS CellSeg to assess the generalization of our method across different imaging conditions and biological objects.

\noindent\textit{Xenium} \cite{pang2025cellotype} contains transcript-based images generated using a spatial transcriptomics platform, 10x Genomics Xenium. Imaging-based spatial transcriptomics data represents another major type of spatial omics data, and is not covered during training. We adopt the test set of Xenium to evaluate our method's robustness when encountered with new imaging equipment (platform).

\noindent\textit{DeepBacs} \cite{spahn2022deepbacs} contains microscopy images of various bacteria species captured under different imaging conditions (E. coli and S. aureus with brightfield microscopy and  S. aureus and B. subtilis with fluorescence microscopy). As bacteria constitute an important class of biological objects in biomedical research and exhibit very distinct morphological and imaging characteristics, we adopt the test set of DeepBacs to assess our method's generalization capability.

\noindent\textit{TEM Bench} \cite{conrad2023instance} contains transmission electron microscopy (TEM) images collected from various sources. Each image is annotated with instance segmentation masks of mitochondria. It is adopted to evaluate our method's performance when encountering new microscopy equipment (TEM) unseen during training.

\noindent\textbf{Tracking datasets}

\noindent\textit{BF-C2DL-MuSC} \cite{mavska2023cell} contains time-lapse microscopy images of mouse muscle stem cells in hydrogel microwells, captured with brightfield microscopy. This dataset is used both for training and evaluation. We adopt the image sequences with publicly available ground-truth annotations in our experiments, considering the 01 set as training set and 02 set as evaluation set, following \cite{scherr2020cell} (the same setting is applied for other tracking datasets).

\noindent\textit{BF-C2DL-HSC} \cite{mavska2023cell} consists of brightfield time-lapse microscopy images of mouse hematopoietic stem cells in hydrogel microwells. This dataset is used both for training and evaluation.

\noindent\textit{Fluo-C2DL-MSC}
\cite{mavska2023cell} contains fluorescence time-lapse microscopy images of rat mesenchymal stem cells on a flat polyacrylamide substrate. This dataset is used both for training and evaluation.

\noindent\textit{PhC-C2DH-U373}
\cite{mavska2023cell} contains phase-contrast time-lapse microscopy images of glioblastoma–astrocytoma U373 cells (derived from a malignant tumor) cultured on a polyacrylamide substrate. This dataset is used both for training and evaluation.

\noindent\textit{DIC-C2DH-HeLa.}
\cite{mavska2023cell} contains time-lapse images of HeLa cells cultured on a flat glass, captured using differential interference contrast microscopy. We use its evaluation set to evaluate the model's generalization ability to a new microscopy modality unseen during framework training. 

\noindent\textit{Fluo-N2DH-GOWT1}
\cite{mavska2023cell,bartova2011recruitment} consists of time-lapse fluorescence microscopy images of GOWT1 mouse stem cells, processed using GFP labeling. We use its evaluation set to evaluate the model's generalization ability when facing new combinations of imaging techniques and biological objects. 

\noindent\textit{Fluo-N2DL-HeLa}
\cite{mavska2023cell,neumann2010phenotypic} contains time-lapse fluorescence microscopy images of HeLa cells stably expressing the nuclear marker H2B-GFP, representing another distinct combination of imaging conditions and biological objects used for evaluation. 

\noindent\textit{Fluo-N2DH-SIM+}
\cite{mavska2023cell} contains simulated fluorescence time-lapse image sequences of HL60 cell nuclei stained with Hoechst, generated using MitoGen  \cite{svoboda2016mitogen}, adopted as distributionally distinct microscopy data for evaluation.

\noindent\textbf{Counting datasets}

\noindent\textit{DCC}
\cite{marsden2018people} consists of brightfield microscopy images of tissues cells, including embryonic mice stem cells, human lung adenocarcinoma, and human monocytes. Each image is provided with dot annotations of the center locations of each cell. This dataset is used for both training and evaluation in our framework.

\noindent\textit{VGG}
\cite{lempitsky2010learning} contains simulated fluorescence microscopy images, with real-life effects such as overlapping cells, cells with shape variability, and imaging artifacts like vignetting. Each image is provided with dot annotations of the center locations of each cell. This dataset is used for both training and evaluation in our framework.

\noindent\textit{ADI}
\cite{paul2017count} consists of images sampled from high-resolution histology slides of human subcutaneous adipose tissue cells. The adipocytes in ADI exhibit distinctive morphological characteristics, such as cytoplasm filled with lipid droplets and large variations in cell size, representing a test case of evaluating model generalization to histological images with distinct cell morphologies. Each image is provided with point annotations of the center locations of the cells.

\noindent\textit{MBM}
\cite{kainz2015you} contains histopathology microscopy images of healthy human bone marrow from eight different patients, where the standard staining protocol depicts cell nuclei in blue and other constituents in shades of pink and red. The dot annotations of each cell are provided for each image. We use this dataset to evaluate the model's generalization ability to H\&E-stained histopathology data.

\subsection*{Compared methods}

\noindent\textit{Cellpose3.} 
Cellpose3 was run using the pre-trained model (Cyto3) fine-tuned on the training sets of Cellpose Cyto, TissueNet, LiveCell, and Lucchi datasets. The fine-tuning was done following the protocol provided by the authors. Key parameter settings include learning rate of 0.1, weight decay of $1e-4$, and max epoch of 100.

\noindent\textit{CelloType.} 
CelloType was run using the pre-trained model (provided by the authors) fine-tuned on the training sets of Cellpose Cyto, TissueNet, LiveCell, and Lucchi datasets. The fine-tuning was done following the protocol provided by the authors. Key parameter settings include learning rate of 0.0001 and max iteration of 5000.

\noindent\textit{MicroSAM.}
MicroSAM was run using the pre-trained model (vit-l-lm) fine-tuned on the training sets of Cellpose Cyto, TissueNet, LiveCell, and Lucchi datasets. The fine-tuning was done following the protocol provided by the authors. Key parameter settings include learning rate of $1e-5$ and max epoch of 100.

\noindent\textit{MitoNet.}
MitoNet was run using the pre-trained model MitoNet\_v1 (provided by the authors) fine-tuned on the training sets of Cellpose Cyto, TissueNet, LiveCell, and Lucchi datasets. The fine-tuning was done following the protocol provided by the authors. Specifically, we employed the Napari plugin for 2D segmentation using the MitoNet\_v1 model with default model and training parameters. All input images were preprocessed into single-channel grayscale format to accommodate MitoNet’s input requirements.

\noindent\textit{TrackMate.}
TrackMate was run using the segmentation results predicted by Cellpose3, a widely adopted segmentation model frequently employed in microscopy tracking workflows \cite{ershov2022trackmate,bragantini2025ultrack}. We run TrackMate with Cellpose3 fine-tuned on the training sets of BF-C2DL-MuSC, BF-C2DL-HSC, Fluo-C2DL-MSC, and PhC-C2DH-U373, following the protocol provided by the authors, with learning rate of 0.0002 and max iteration of 500.

\noindent\textit{Trackastra.}
Trackastra was run using the segmentation results predicted by Cellpose3. We run Trackastra with Cellpose3 fine-tuned on the training sets of BF-C2DL-MuSC, BF-C2DL-HSC, Fluo-C2DL-MSC, and PhC-C2DH-U373, with learning rate of 0.0002 and max iteration of 500.

\noindent\textit{Ultrack.}
Ultrack was run using the segmentation results predicted by Cellpose3. We run Ultrack with Cellpose3 fine-tuned on the training sets of BF-C2DL-MuSC, BF-C2DL-HSC, Fluo-C2DL-MSC, and PhC-C2DH-U373, with learning rate of 0.0002 and max iteration of 500. 

\noindent\textit{U-Net.}
We run U-Net trained on the training sets of DCC and VGG, following the training protocol in \cite{falk2019u,guo2019sau}, with learning rate of 0.001 and max epoch of 300.

\noindent\textit{DCL.}
We run DCL trained on the training sets of DCC and VGG, following the training protocol in \cite{zheng2024rethinking}, with learning rate of $1e-4$ and max epoch of 300.

\subsection*{Details of the collection of image data from diverse, real-world experiments}

For this evaluation, we collect microscopy image data (images and image sequences) from publicly available resources, including the Image Data Resource~\cite{williams2017image}, the Cell Image Library~\cite{CellImageLibrary}, and the Bioimage Archive~\cite{hartley2022bioimage}. Each image sample is captured under a distinct microscopy image analysis setting, varying in subject species, biological objects, microscopy equipment, sample preparation protocols, analysis goals, and research facilities. The resulting collection corresponds to a total of 200 sets of distinct experiments, spanning a broad range of biological contexts, including different species (such as human, freshwater ciliates, microalgae, parasitic protozoa, flowering plants, bacteria, etc.), biological targets (e.g., mammalian cells such as umbilical vein endothelial cells and embryonic stem cells, cells obtained from clinical biopsy specimens, and subcellular structures such as azurophil granules, etc.), staining protocols (e.g., hematoxylin and eosin, DAPI, immunofluorescence, alkaline phosphatase staining, etc.), and microscopy conditions (e.g., brightfield microscopy, widefield fluorescence microscopy, confocal microscopy, scanning electron microscopy, etc.). Together, this collection reflects substantial real-world heterogeneity. The information for each individual test sample is provided in Supplementary Table 13. The collected image data were annotated by five practitioners, who independently generated annotations and then cross-checked and reconciled any discrepancies. The collected image data and the annotations are made publicly available at https://github.com/phoebehxf/MicroscopyMatching.

\subsection*{Details of expert rating}

To assess the perceived quality of MicroscopyMatching's predictions, we conducted an expert rating evaluation involving eight biomedical professionals. For each of the three tasks (segmentation, tracking, and counting), a subset of input image samples (images or image sequences) was randomly selected from the test sets of the corresponding benchmark datasets. For each input image sample, each professional was presented with the input image data itself, together with two corresponding outputs: one generated by MicroscopyMatching and the other obtained via human annotation. For clearer visualization, outputs were visualized in a task-specific manner: for segmentation tasks, the outputs were presented by overlaying the generated masks on the input image; for tracking tasks, trajectories were shown on top of time-lapse sequences; and for counting tasks, the generated density map together with the resulting object count were displayed.

For each image sample, the two results (i.e., the human annotation and the MicroscopyMatching prediction) were presented in random order, with the former in the random order labeled as ``result1'' and the latter as ``result2''. Each professional then performed a blinded comparison between the two results and assigned a score on a 5-point Likert scale reflecting their relative quality (1 = ``result1'' is clearly superior; 2 = ``result1'' is superior; 3 = ``result1'' is of similar quality to ``result2''; 4 = ``result1'' is inferior; 5 = ``result1'' is clearly inferior). After aggregating the expert ratings, we converted each score into a relative performance score for MicroscopyMatching ranging from -2 to +2, where +2 indicates that MicroscopyMatching's prediction is much better than the human annotation and -2 indicates the opposite. In total, 2,000 pairwise comparisons were collected.

\subsection*{Details of human-model agreement evaluation}

To more meticulously assess MicroscopyMatching’s performance, we also conducted a human-model agreement evaluation. Specifically, we provided each of five annotators (members of a digital health laboratory) with the same set of microscopy image samples (images and image sequences) collected from diverse, real-world experimental settings and asked them to independently annotate each sample. We quantified inter-annotator variability and compared it with MicroscopyMatching’s agreement with each annotator. In our evaluation, we used mean AP as the metric for segmentation, TRA for tracking, and MAE for counting. Across tasks, MicroscopyMatching’s agreement with individual annotators (human-model agreement) falls within the range of inter-annotator (human-human) variability, indicating that MicroscopyMatching’s performance is comparable to that of human annotators.

\subsection*{Details of user study}

For the user study, we invited ten biomedical researchers and practitioners from different backgrounds and regions (including Australia, China, the United Kingdom, and Singapore). Specifically, these participants included: three clinical laboratory physicians working in hospitals (with two of them routinely performing microscopic analyses of samples such as blood, urine, and other bodily fluids, and the other conducting immunofluorescence test), two biological researchers (one of which specializing in birds and the other in plants), two medical graduate students (one studying tumors in brain tissues and the other studying bone-related diseases), and three biology student (one studying microorganism such as bacteria, one specializing in confocal microscopy, and the other studying brightfield microscopy).

We assessed the practical utility of MicroscopyMatching in the user study from two complementary perspectives: (1) objective performance, by comparing MicroscopyMatching's performance with manual human annotations; and (2) subjective user experience, by collecting participants' feedback after their hands-on use with MicroscopyMatching. In the user study, participants freely chose whether to provide a bounding-box exemplar (corresponding to the MicroscopyMatching\_S mode) or rely solely on semantic matching without an exemplar (MicroscopyMatching\_A mode). For each task, the prediction performance, the task completion time, and bounding-box exemplar usage were recorded (81\% of the tasks were completed without using exemplar bounding boxes). To assess MicroscopyMatching's objective performance and its practical utility as a reliable and ready-to-use alternative to manual annotation, we compared its performance and task completion time directly with human annotations. Specifically, both MicroscopyMatching's outputs and human annotations were evaluated using ground-truth annotations, which were established through consensus review by two independent experts. On the other hand, for subjective user experience evaluation, participants were asked to complete a post-task questionnaire and rated their experience using MicroscopyMatching (including ease of use, the interpretability of the predicted results, their satisfaction with the performance, and the perceived readiness of MicroscopyMatching) on a five-point Likert scale. Together, these evaluations provide a comprehensive assessment of MicroscopyMatching's practical utility in real-world microscopy analysis scenarios for biomedical researchers and practitioners.

\subsection*{Details of practical guidelines}
In the first exploration in the ``Practical Guidelines for Providing Exemplars'' part of the Results section, we investigated the impact of using different number of bounding boxes as input for MicroscopyMatching. In this exploration, the bounding boxes are randomly selected and derived from the object annotations. We then elaborate on how MicroscopyMatching seamlessly supports multiple bounding boxes as input. Specifically, given the input image and $m$ bounding boxes when $m > 1$ (denoted as  $\{b_1, b_2, ..., b_m\}$), we pass the input image and the bounding boxes into the exemplar projector to obtain $m$ corresponding conditioning embeddings. We average these embeddings to obtain the final conditioning embedding $e$. The subsequent process follows the same pipeline as using one exemplar bounding box.

In the second exploration, we investigated the impact of using imprecisely specified exemplar bounding boxes in MicroscopyMatching. We follow the taxonomy of different types of bounding box imprecision in \cite{bolya2020tide,strafforello2022humans,zheng2020distance}. Specifically, these include shift error, size error, and aspect ratio error. A bounding box is regarded as erroneous (imprecise) when its IoU with the ground-truth box falls below 0.5. An erroneous bounding box is categorized as a shift error if its center is displaced by more than 50\% of the ground-truth box’s width or height. A size error is defined when the annotated box differs in scale from the ground truth by more than 50\% in area. An aspect ratio error occurs when the ratio between the box's width and height deviates by more than 50\% from that of the ground truth. The experiments were conducted using microscopy image samples collected across diverse real-world experiments.

\end{document}